\documentclass[runningheads]{llncs}

\usepackage{eccv}

\usepackage{eccvabbrv}

\usepackage{graphicx}
\usepackage{booktabs}
\usepackage{pifont} 
\usepackage{bm}
\usepackage[table]{xcolor}

\definecolor{GrayXMark}{gray}{0.5}
\newcommand{\xmark}{ {\color{gray} \ding{55}} }
\definecolor{myblue}{rgb}{0.05,0.3,0.78}
\definecolor{mygreen}{rgb}{0.2,0.5,0.11}
\definecolor{tbd}{rgb}{1.0, 0.0, 0.5}

\usepackage{hyperref}

\usepackage{orcidlink}
\newcommand{\SM}[1]{{\color{red}{#1}}}

\begin{document}
\title{Formula-Supervised \\ Visual-Geometric Pre-training} 


\author{Ryosuke Yamada*\inst{1,2}\orcidlink{0000-0002-2154-8230} \and
Kensho Hara*\inst{1}\orcidlink{0000-0001-6463-7738} \and \\
Hirokatsu Kataoka\inst{1}\orcidlink{0000-0001-8844-165X} \and
Koshi Makihara\inst{1}\orcidlink{0000-0003-4145-7595} \and
Nakamasa Inoue\inst{1,3}\orcidlink{0000-0002-9761-4142} \and \\
Rio Yokota\inst{1,3}\orcidlink{0000-0001-7573-7873} \and
Yutaka Satoh\inst{1,2}\orcidlink{0000-0002-0638-0855}
}

\authorrunning{R.~Yamada et al.}

\institute{
  1 AIST, \quad 2 University of Tsukuba, \quad 3 Tokyo Institute of Technology \\
  * equal contribution \\
  \vspace{2pt}
  Project Page: \url{https://ryosuke-yamada.github.io/fdsl-fsvgp/}
}

\maketitle

\begin{abstract}
Throughout the history of computer vision, while research has explored the integration of images (visual) and point clouds (geometric), many advancements in image and 3D object recognition have tended to process these modalities separately.
We aim to bridge this divide by integrating images and point clouds on a unified transformer model. This approach integrates the modality-specific properties of images and point clouds and achieves fundamental downstream tasks in image and 3D object recognition on a unified transformer model by learning visual-geometric representations. In this work, we introduce \textbf{F}ormula-\textbf{S}upervised \textbf{V}isual-\textbf{G}eometric \textbf{P}re-training (\textbf{FSVGP}), a novel synthetic pre-training method that automatically generates aligned synthetic images and point clouds from mathematical formulas. Through cross-modality supervision, we enable supervised pre-training between visual and geometric modalities. FSVGP also reduces reliance on real data collection, cross-modality alignment, and human annotation. Our experimental results show that FSVGP pre-trains more effectively than VisualAtom and PC-FractalDB across six tasks: image and 3D object classification, detection, and segmentation. These achievements demonstrate FSVGP's superior generalization in image and 3D object recognition and underscore the potential of synthetic pre-training in visual-geometric representation learning. Our project website is available at \url{https://ryosuke-yamada.github.io/fdsl-fsvgp/}.


\end{abstract}    
\section{Introduction}
\label{sec:intro}
Fusing images (visual) and point clouds (geometric) is crucial for developing vision models that enhance understanding of the real world. This is because the visual and geometric modalities are complementary.
For example, a vision model that relies only on point clouds cannot distinguish a picture and a poster attached to the wall. However, the difference in texture between these two objects can easily be identified by fusing images in the same 3D scene.
Therefore, extracting visual-geometric representations by integrating images and point clouds enhances the recognition capabilities of vision models.

\begin{figure}[t]
    \centering
    \includegraphics[width=1.0\linewidth]{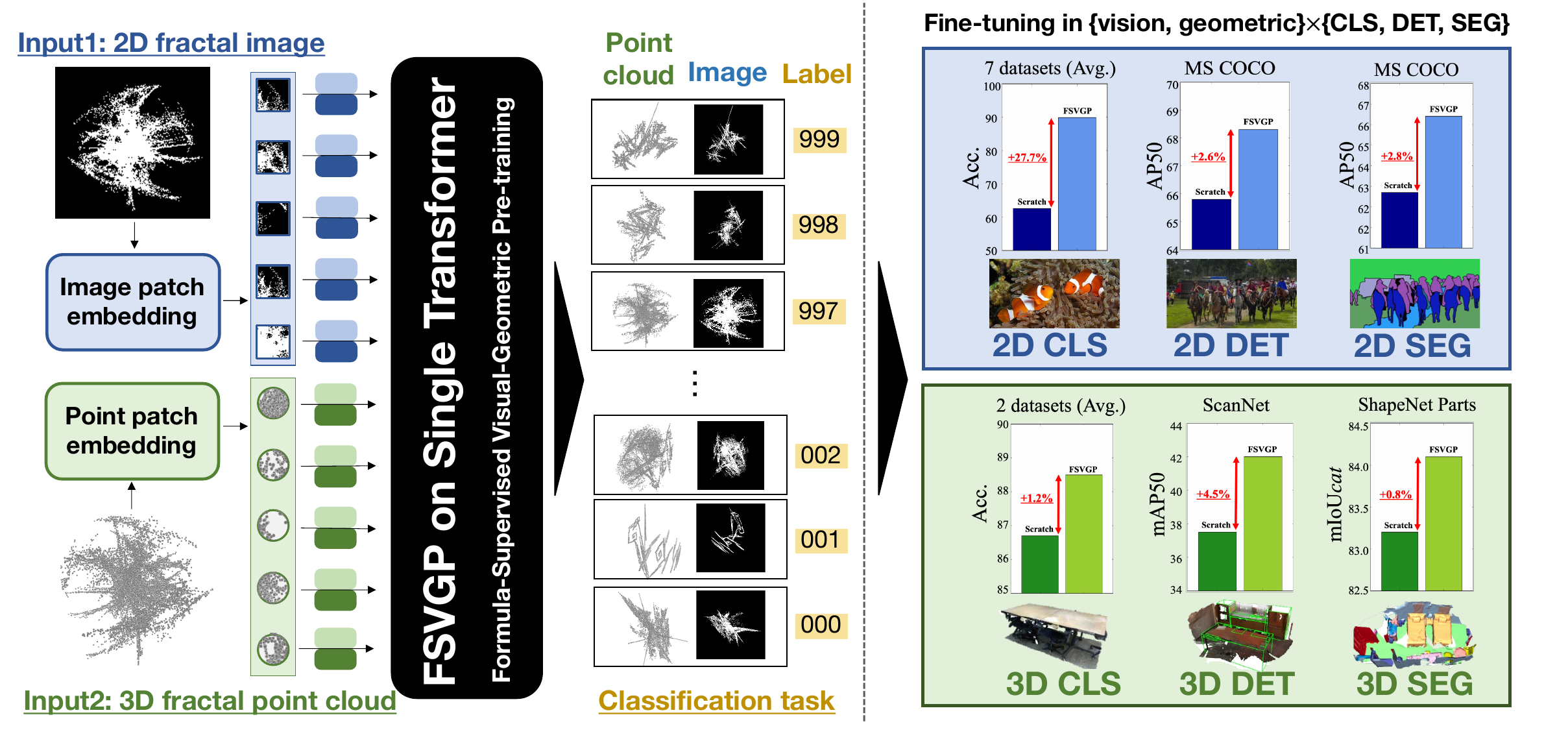}
    \vspace{-10pt}
    \caption{
    FSVGP enables pre-training visual and geometric modalities on a unified transformer model by constructing VG-FractalDB from a mathematical formula. 
    VG-FractalDB consists of fractal images, fractal point clouds, and cross-modal supervision called formula-supervised consistency labels. FSVGP simultaneously inputs a fractal image and a fractal point cloud and pre-trains in classification (CLS) tasks based on a formula-supervised consistency label. 
    We show that FSVGP improves six tasks of image and 3D object CLS, detection (DET), and segmentation (SEG).
    }
    \label{fig1}
    \vspace{-20pt}
\end{figure}

Despite the ongoing research in visual-geometric representation learning that utilizes images and point clouds, a significant gap exists in developing a unified vision model that effectively trains on both modalities, enhancing both image and 3D object recognition capabilities. 
Within the realm of visual-geometric representation learning, various studies~\cite{hou2021pri3d,wang2022p2p,liu20213d,kundu2020virtual} have pursued improvements in image recognition by integrating visual and geometric data, while others~\cite{afham2022crosspoint,robert2022learning,liu2021learning,yi2023invariant,tang2023prototransfer} have aimed to enhance 3D object recognition. In 2024, recognition models have emerged that are limited to segmentation tasks yet can address both image and 3D data~\cite{jain2024odin}.
The challenge is partly due to the scarcity of large-scale datasets pairing images with point clouds, suggesting extensive paired data is necessary to bridge visual and geometric modalities effectively.

However, given the scarcity of high-quality 3D data on the web, collecting paired images and point clouds proves significantly more challenging and costly.
Furthermore, accurate annotations often necessitate manual labeling by experts who can interpret complex spatial information of 3D data. 
In addition, aligning images with point clouds requires significant correspondence and pre-processing costs to address distortions in raw point clouds and their associated image projections, which are prone to distortion.
Consequently, the construction of large-scale datasets for visual and geometric modalities poses a formidable challenge, demanding substantial human resources and specialized expertise.
Moreover, copyright and ethical biases are becoming an increasing concern on real datasets.

We facilitate visual-geometric representation learning by employing formula-driven supervised learning (FDSL) to address these challenges. FDSL~\cite{kataoka2020pre} automatically generates synthetic data and supervision from a mathematical formula based on principles such as fractal geometry. Furthermore, FDSL helps circumvent common issues associated with real data, including manual labeling costs, copyright concerns, and ethical biases.

Hence, we introduce a visual-geometric pre-training method called \textbf{F}ormula-\textbf{S}upervised \textbf{V}isual-\textbf{G}eometric \textbf{P}re-training (\textbf{FSVGP}). 
FSVGP enables synthetic pre-training through a unified transformer model by automatically generating aligned synthetic images and point clouds, as shown in Figure~\ref{fig1}. 
We first developed the visual-geometric fractal database (VG-FractalDB), which employs fractal geometry to generate fractal point clouds and their corresponding fractal images automatically, processed simultaneously by a unified transformer model.
VG-FractalDB provides a formula-supervised consistency label as cross-modality supervision between visual and geometric modalities.
The formula-supervised consistency labels ensure correspondences between fractal images and fractal point clouds, facilitating supervised pre-training for classification tasks on a unified transformer model.
For the transformer model, we made minimal modifications to the input processing—drawing upon the Vision Transformer (ViT)~\cite{dosovitskiy2021an} and Point Transformer (PointT)~\cite{zhao2021point}—to maintain its flexibility.
Thus, FSVGP achieves synthetic pre-training, effectively learning visual-geometric representations to integrate images and point clouds into a unified transformer model. This facilitates image and 3D object recognition using a unified transformer model.

In summary, FSVGP is a novel supervised synthetic pre-training designed to train the VG-FractalDB on a unified transformer model. Our contributions are as follows: (i) Our experimental results show that FSVGP improves fine-tuning performance across six tasks, including image and 3D object classification, detection, and segmentation. (ii) We demonstrate that FSVGP surpasses the latest FDSL method (VisualAtom) in image classification, detection, and segmentation tasks. (iii)  We show that FSVGP is superior to the latest FDSL method (PC-FractalDB) in 3D object classification, detection, and segmentation tasks.

\section{Related work}
\label{sec:related_work}
In this section, we limit the discussion of related work to that closely related to the proposed FSVGP.

In image recognition, recent self-supervised learning (SSL) methods~\cite{dehghani2023scaling,Caron_2021_ICCV,he2022masked} using massive datasets such as JFT-300M~\cite{sun2017revisiting} or ImageNet-21k have begun to surpass the longstanding de facto standard of ImageNet-1k for supervised pre-training. 
Various SSL methods ~\cite{Yu_2022_CVPR,pang2022masked,liu2022masked} in 3D object recognition, utilizing ShapeNet~\cite{chang2015shapenet}, have been proposed, showcasing certain pre-training effects in downstream tasks. 
In addition, large-scale 3D datasets such as Objaverse-XL~\cite{deitke2023objaversexl} have recently been introduced in 3D vision tasks.
Nevertheless, using large-scale real datasets raises ethical concerns, including manual labeling costs, copyright issues, and biases.
It is possible to delete 3D objects from the web by their creators in Objaverse-XL. 
The importance of open-source datasets becomes evident as models like ViT-22B~\cite{dehghani2023scaling} are often trained with non-public datasets, underscoring the need for transparency and accessibility in computer vision.

Visual-geometric representation learning with images and point clouds aims to improve recognition performance over a single modality vision model. This research is categorized into two main areas: enhancing visual and geometric recognition. For the former, the goal is to integrate geometric information from 3D data into visual representations to enrich the understanding of 3D scenes~\cite{hou2021pri3d,wang2022p2p,liu20213d,kundu2020virtual}.
For instance, Pri3D~\cite{hou2021pri3d} utilizes contrastive learning to fuse the relationship between corresponding point clouds and pixels.
Conversely, methods to enhance 3D object  recognition leverage visual knowledge derived from massive images~\cite{afham2022crosspoint,robert2022learning,liu2021learning,yi2023invariant,tang2023prototransfer}. 
CrossPoint~\cite{afham2022crosspoint} uses contrastive learning between point clouds and multi-view images.

Furthermore, FDSL is a notable method that addresses pre-training real dataset limitations~\cite{kataoka2020pre,Kataoka_2022_CVPR,takashima2023visual,yamada2022point,shinoda2023segrcdb,tadokoro2024primitive,chiche2024pre}. 
Unlike SSL, which assigns pseudo-labels to unlabeled data, FDSL uses a mathematical formula to generate synthetic data and corresponding labels for pre-training. 
For instance, VisualAtom~\cite{takashima2023visual} utilizes FDSL to generate synthetic images with complex contours to pre-train ViT, demonstrating effectiveness in image classification, detection, and segmentation. Similarly, PC-FractalDB~\cite{yamada2022point}, a synthetic 3D scene dataset, enhances fine-tuning performance in 3D object detection through VoteNet~\cite{qi2019deep} pre-training.
However, the previous FDSL mainly focused on specific modalities and tasks.

Thus, we introduce FSVGP, which extends FDSL to visual-geometric representation learning, achieving supervised pre-training on a unified transformer model. 
FSVGP can effectively serve as a backbone network for a broad image and 3D object recognition spectrum through we develop vanilla transformers with minimal modification.
Moreover, since FSVGP utilizes synthetic data, it circumvents the ethical issues of real data.
\section{Formula-supervised visual-geometric pre-training}
\vspace{-10pt}
This section introduces FSVGP, a novel synthetic pre-training method designed to learn visual-geometric representations for image and 3D object recognition.
Unlike previous visual-geometric representation learning methods that predominantly focus on image or 3D object recognition in isolation, FSVGP trains visual-geometric representations on a unified transformer model.

\begin{figure}[t]
  \centering
 \includegraphics[width=0.95\linewidth]{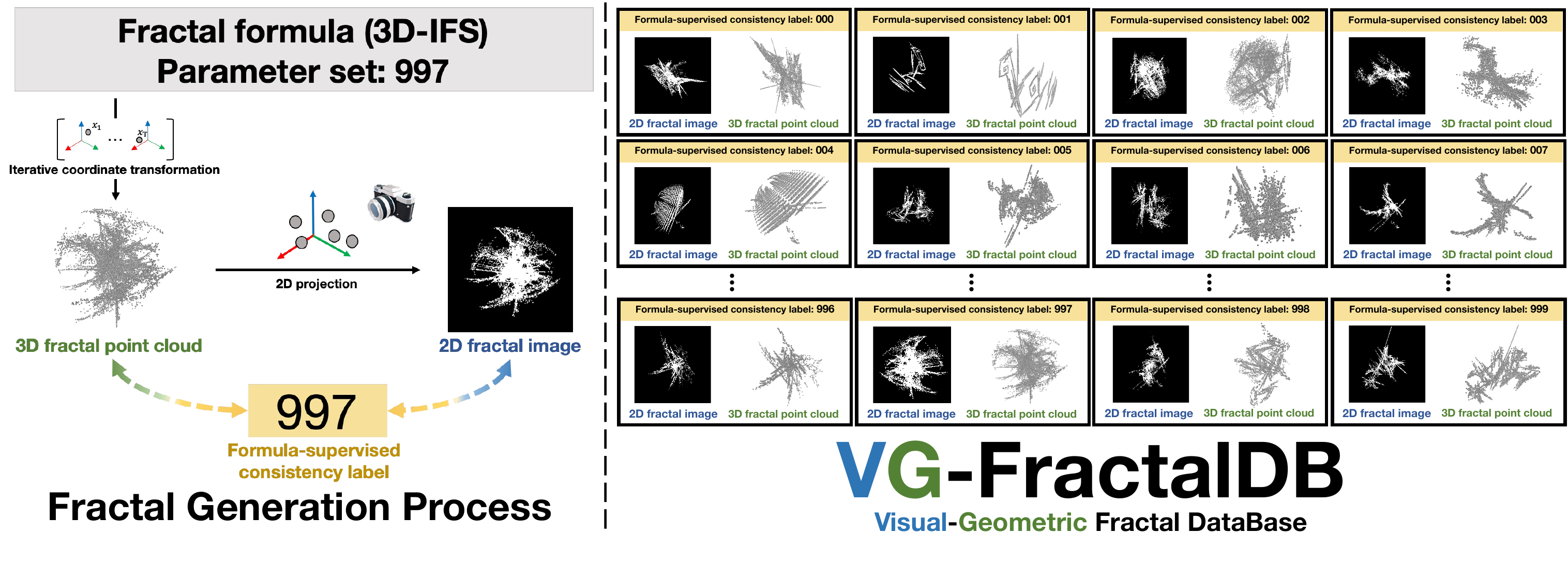}
 \vspace{-5pt}
 \caption{
 {\bf Overview of the fractal generation process and VG-FractalDB.} The fractal generation process creates paired fractal data and formula-supervised consistency labels. Initially, fractal point clouds are generated using the 3D Iterated Function System (3D-IFS). The fractal point clouds are then projected onto 2D planes to form fractal images. Simultaneously, formula-supervised consistency labels are automatically generated based on the variance of 3D coordinates, serving as cross-modality supervision. 
 We construct the VG-FractalDB by repeating these generations.
 }
\vspace{-10pt}
\label{fig:vgfractaldb}
\end{figure}

To implement FSVGP, we construct VG-FractalDB, which automatically generates fractal images and fractal point clouds based on fractal geometry, as shown in Figure~\ref{fig:vgfractaldb}.
The generation process of fractal images and fractal point clouds refers to previous research~\cite{yamada2021mv,yamada2022point}. 
Our study's key distinction from previous research~\cite{yamada2021mv,yamada2022point} is the implementation of supervised pre-training on a unified transformer model by utilizing formula-supervised consistency labels between visual and geometric modalities derived directly from a mathematical formula as cross-modality supervision. 
As a result, our approach supports visual-geometric learning within a shared label space.
Furthermore, our study's concept is based on the fact that FSVGP learns the natural law of visual-geometric relationships by using fractal geometry as a generation rule.

Furthermore, we do not develop special cross-modal modules and complex multi-task learning. As a result, by implementing minimal modifications to the transformer model and employing a straightforward loss function for pre-training, we not only facilitate visual-geometric representation learning but also broaden the FSVGP's range of applicability.

\vspace{-10pt}
\subsection{Visual-geometric fractal database (VG-FractalDB)}
\label{sec:VG-FractalDB}
VG-FractalDB is a pre-training dataset that consists of fractal images and fractal point clouds with formula-supervised consistency labels (see Figure~\ref{fig:vgfractaldb}).
Specifically, the VG-FractalDB is defined by $\mathcal{D} = \{({X}_{j}, I_{j}, y_{j})\}_{j=1}^{N}$, where $X_{j}$ represents a fractal point cloud, $I_{j}$ indicates a fractal image, and $y_{j}$ is a formula-supervised consistency label that relates to both visual and geometric modalities. 
Here, $N$ signifies the total number of pre-training data. The formula-supervised consistency labels are categorized discretely within $\{1, 2, \cdots, C\}$, where $C$ denotes the number of categories. Note that the number of data in VG-FractalDB is $N$, since the fractal image is not generated by the other formulas but by simply projecting a fractal point cloud onto a 2D image plane.

\noindent{\textbf{Geometric modality -- fractal point cloud.}}
A fractal point cloud is generated using the 3D Iterated Function System (3D-IFS), a method rooted in fractal geometry for creating complex, self-similar structures. The 3D-IFS for generating a fractal point cloud of category $c$, denoted as $\Theta^c$, is defined as $\Theta^c = \{\mathcal{X}; t_{1}^c, t_{2}^c, \cdots, t_{n}^c; p_{1}^c, p_{2}^c, \cdots, p_{n}^c\}$, where each $t_{i}^c : \mathcal{X} \to \mathcal{X}$ represents an affine transformation function within the space $\mathcal{X}$, and $\{p_{i}^c\}$ are the associated probabilities of each transformation, with $n=7$ denoting the number of transformations.
A fractal point cloud is represented as a set of coordinates, $\{\bm{x}_{t}\}_{t=1}^{T}$, within the complete metric space $\mathcal{X}$, here employing the 3D Euclidean space $\mathcal{X} = \mathbb{R}^{3}$. The point cloud is generated by a series of affine transformations $t_{i}$ applied within this space.
Each affine transformation $t_{i}$ is defined by the equation
$t_{i}(\bm{x}) = \bm{r}_{i} \bm{x} + \bm{b}_{i}$,
where $\bm{r}_{i}$ is a transformation matrix within $\mathbb{R}^{3 \times 3}$, $\bm{x}$ represents the coordinates within $\mathbb{R}^{3}$, and $\bm{b}_{i}$ is a bias vector. The initial point $\bm{x}_{1}$ is conventionally set to the origin (zero vector).
The number of affine transformations, $n$, along with the elements of each rotation matrix $\bm{r}_{i}$ and bias vector $\bm{b}_{i}$, are determined through random sampling from specified uniform distributions. The transformation probability, $p_{i}$, is calculated proportionally to the determinant of $\bm{r}_{i}$, normalized by the sum of determinants for all transformations.
The final coordinate set, $X$, comprises the sequence of coordinates generated up to a predetermined limit $T=8192$.

\noindent \textbf{Visual modality -- fractal image.}
We detail transforming a fractal point cloud into a fractal image by projecting it onto an image plane. This conversion employs a mapping, $\mathcal{F}_{\text{RGB}}$, which interprets the coordinates in $X_{j}$ as white dots against a black background, effectively rendering the fractal geometry visually in two dimensions. To facilitate this transformation, we introduce a virtual camera, $\bm{c}$, designed to map the 3D coordinate set into a fractal image, $I_{j}$. The mapping process is succinctly represented as $I_{j} = \mathcal{F}_{\text{RGB}}(X_{j}; \bm{c})$.

To ensure a precise alignment between the fractal images and their corresponding fractal point clouds—thus maintaining a coherent pairing of visual and geometric data—we allocate one virtual camera, $\bm{c}_{v}$, per fractal point cloud, where $v = 1$. The chosen method for projection is perspective projection, offering a realistic spatial representation. The positioning of the virtual camera is determined randomly but is strategically placed on a sphere that centers around the fractal object's center of gravity, optimizing the view of the fractal's intricate structures.
The fractal image size is set to $(W, H) = (224, 224)$.

\noindent{\textbf{Formula-supervised consistency label.}}
We introduce a new approach that assigns formula-based supervision, termed ``formula-supervised consistency labels,'' to fractal images and fractal point clouds, originating from a unified mathematical formula (see Figure~\ref{fig:vgfractaldb} Left). These labels emerge from mathematical formulas, enabling the definition of common labels across modalities—a process traditionally requiring costly and specialized pre-processing to map images to point clouds and vice versa. Moreover, formula-supervised consistency labels facilitate the simultaneous input of fractal images and fractal point clouds into a unified transformer model, promoting learning within a shared label space.

\begin{figure}[t]
  \centering
 \includegraphics[width=0.95\linewidth]{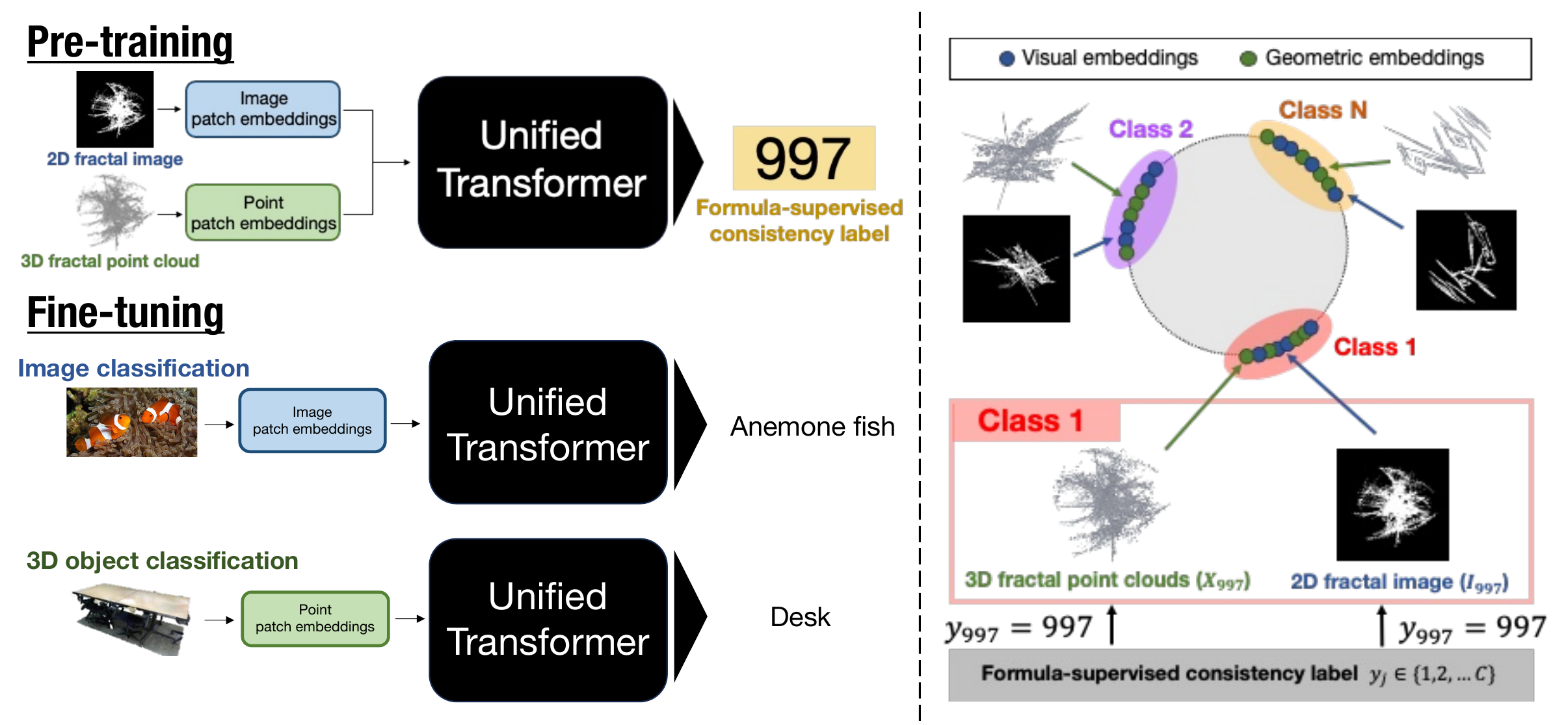}
 \vspace{-5pt}
 \caption{\textbf{VG-FractalDB pre-training.} \textbf{Left:} We trains VG-FractalDB on a unified transformer model. After pre-training, we can fine-tune the image and 3D object recognition by using the same unified transformer model. \textbf{Right:} FSVGP learns visual and geometric modalities by supervised pre-training based on a formula-supervised consistency label. Therefore, FSVGP can train different modalities within a common label space on a unified transformer model.
 }
\vspace{-10pt}
\label{fig:overview}
\end{figure}

As described above, a fractal category $c$ is defined by 3D-IFS $\Theta^c$. A 3D fractal point cloud is generated by the 3D-IFS and is projected onto a 2D image plane to generate a 2D fractal image. Therefore, the 2D fractal image and 3D fractal point cloud share the consistent label.
To remove ineffective fractal categories, we employ a variance threshold criterion. 
This algorithm assesses whether the variance of a fractal category exceeds a predefined threshold (0.05) along each coordinate axis. We determined the threshold with reference to~\cite{yamada2022point}. 
In other words, we only define fractal point clouds that are above the threshold as a fractal category.
To enrich the diversity of instances within each fractal category, we implement a technique named FractalNoiseMix, as described in~\cite{yamada2022point}. This method integrates an additional 20\% of points generated from randomly selected fractal point clouds in VG-FractalDB, enhancing the dataset's robustness and variability.
Further details on VG-FractalDB, including examples and parameter specifications, are available in the \SM{Supplementary Material}.

\subsection{VG-FractalDB Pre-training on a unified transformer model}
\vspace{-5pt}
\label{sec:vgfractal_pretrained_vit}
To train VG-FractalDB with a unified transformer model, we modified the ViT and PointT to input a fractal image and a fractal point cloud simultaneously, as shown in Figure~\ref{fig:overview} Left. Our modifications were limited to the input processing, ensuring the transformer model remains as straightforward as possible. This approach maintains the model's adaptability to various tasks without compromising its pre-training flexibility.

For embedding vectors specific to each modality, we utilized ViT for fractal images and PointT for fractal point clouds. Specifically, a fractal image and a fractal point cloud are divided and embedded into the image tokens $\mathbf{z}_\mathrm{i} = [ x_\mathrm{class}, \mathbf{z}_\mathrm{i}^1, \mathbf{z}_\mathrm{i}^2, \dots, \mathbf{z}_\mathrm{i}^{M_\mathrm{i}} ]$ and point cloud tokens $ \mathbf{z}_\mathrm{p} = [ x_\mathrm{class}, \mathbf{z}_\mathrm{p}^1, \mathbf{z}_\mathrm{p}^2, \dots, \mathbf{z}_\mathrm{p}^{M_\mathrm{p}} ]$, where \( M_\mathrm{i} \) and \( M_\mathrm{p} \) are the numbers of image and point cloud tokens, respectively. 
Moreover, a class token $x_\mathrm{class}$ is added to the tokens of each modality. 
The fractal image and fractal point cloud tokens are then input into the transformer encoder. The class token $x_\mathrm{class}$ and the MLP layer used for classification are shared between the two modalities.

In addition, for the pre-training task of VG-FractalDB, we train our transformer model $f$ using cross-entropy (CE) loss for the classification task (see Figure~\ref{fig:overview} Left), which is given by
\begin{align}
\mathcal{L}_{\text{ce}}(f(\mathcal{D})) = - \frac{1}{N} \sum_{j=1}^{N} \sum_{c=1}^{C} y_{j,c} \log \hat y_{j,c}
\end{align}
where $\hat y_{j} = f(X_{j}, I_{j})$ is the output vector. 
FSVGP trains a unified transformer model on VG-FractalDB to minimize CE loss using AdamW~\cite{loshchilov2017decoupled}.
Our approach to supervised pre-training distinguishes itself from conventional visual-geometric learning methods in several key ways. Whereas the visual-geometric representation learning method~\cite{liu2021learning} focused on pixel-point correspondence for training, FSVGP facilitates learning across different modalities within a common label space (see Figure~\ref{fig:overview} Right). This unified label space allows for optimizing a unified transformer model, the learning process across modalities.

\begin{table*}[t]
\begin{center}
 \caption{Details of fine-tuning datasets, fine-tuning task, model type, number of classes (\#classes), training data (\#train), validation data (\#val), and evaluation metrics.}
 \label{tab:datasets}
 \vspace{-10pt}
 \scalebox{0.74}{
\begin{tabular}{lccccccc}\toprule[0.8pt]
Fine-tuning dataset & Fine-tuning task & Table & Model type &\#classes & \#train & \#val & Metrics \\ \midrule[0.5pt]
\color{myblue}{CIFAR10} (C10)~\cite{Krizhevsky2009_cifar} & \color{myblue}{Image CLS} & \ref{tab:cross-modality_classification}, \ref{tab:comparison_sl_ssl_fdsl} & ViT-B & 10 & 50k & 10k  & Acc. \\ 
\color{myblue}{CIFAR100} (C100)~\cite{Krizhevsky2009_cifar} & \color{myblue}{Image CLS} & \ref{tab:cross-modality_classification}, \ref{tab:comparison_sl_ssl_fdsl} & ViT-B & 100 & 50k & 10k & Acc. \\ 
\color{myblue}{Stanford Cars} (Cars)~\cite{Krause3DRR2013_cars} & \color{myblue}{Image CLS} & \ref{tab:cross-modality_classification}, \ref{tab:comparison_sl_ssl_fdsl}& ViT-B & 196 & 8k & 8k & Acc. \\ 
\color{myblue}{Oxford Flowers} (Flowers)~\cite{Nilsback08_flowers} & \color{myblue}{Image CLS} & \ref{tab:cross-modality_classification}, \ref{tab:comparison_sl_ssl_fdsl}, & ViT-B & 102 & 6k & 818 & Acc. \\ 
\color{myblue}{PascalVOC 2012} (VOC)~\cite{EveringhamIJCV2015_voc} & \color{myblue}{Image CLS} & \ref{tab:cross-modality_classification}, \ref{tab:comparison_sl_ssl_fdsl} & ViT-B & 20 & 13k & 13k & Acc. \\ 
\color{myblue}{Places30} (P30)~\cite{ZhouTPAMI2017_Places} & \color{myblue}{Image CLS} & \ref{tab:cross-modality_classification}, \ref{tab:comparison_sl_ssl_fdsl} & ViT-B & 30 & 150k & 3k & Acc. \\ 
\color{myblue}{ImageNet100} (IN100)~\cite{DengCVPR2009_ImageNet} & \color{myblue}{Image CLS} & \ref{tab:comparison_sixtask}, \ref{tab:cross-modality_classification}, \ref{tab:comparison_sl_ssl_fdsl}, \ref{tab:shapenet}, \ref{tab:perlin}, \ref{tab:pt_task} & ViT-B & 100 & 120k & 5k & Acc. \\ 
\color{myblue}{MS COCO 2017} (COCO)~\cite{LinECCV2014_coco} & \color{myblue}{Image DET / SEG} & \ref{tab:comparison_sixtask}, \ref{tab:detection} & ViTDet-B & 80 & 118k & 5k & AP \\ 
\color{myblue}{ImageNet-1k}~\cite{DengCVPR2009_ImageNet} & \color{myblue}{Image CLS} & \ref{tab:comparison_imagenet1k} & ViT-B & 1000 & 1.2M & 50k & Acc. \\ 
\midrule[0.5pt]
\color{mygreen}{ModelNet40} (M40)~\cite{wu20153d} &  \color{mygreen}{3D object CLS} & \ref{tab:cross-modality_classification}, \ref{tab:3dcls}, \ref{tab:few-shot}, \ref{tab:shapenet}, \ref{tab:perlin}, \ref{tab:pt_task} & PointT-S & 40 & 9.8k & 2.4k & Acc. \\ 
\color{mygreen}{ScanObjectNN} (SONN)~\cite{uy2019revisiting} &  \color{mygreen}{3D object CLS} & \ref{tab:comparison_sixtask}, \ref{tab:cross-modality_classification}, \ref{tab:3dcls}, \ref{tab:few-shot} & PointT-S & 15 & 2.3k & 581 & Acc. \\ 
\color{mygreen}{ShapeNet-Parts}~\cite{uy2019revisiting} &  \color{mygreen}{3D object (parts) SEG} & \ref{tab:comparison_sixtask}, \ref{tab:3ddet} & PointT-S & 15 & 14k & 2.8k & mIoU \\
\color{mygreen}{ScanNet}~\cite{dai2017scannet} & \color{mygreen}{3D object DET} & \ref{tab:comparison_sixtask}, \ref{tab:3ddet} & 3DETR & 18 & 1.2k & 312 & mAP \\
\bottomrule[0.8pt]
\label{tab:datasets}
\end{tabular}}
\end{center}
\vspace{-30pt}
\end{table*}

\section{Experiments}
\vspace{-5pt}
In this section, we evaluate the effectiveness of FSVGP by comparing it with previous pre-training methods. First, Section~\ref{Experimental_setting} outlines our experimental setup. In addition, Section~\ref{brief_result} briefly describes the main experimental results in image recognition and 3D object recognition with FSVGP.  
Subsequently, Sections~\ref{2D_image_recognition} and \ref{3D_object_recognition} compare FSVGP with established pre-training methods across six vision tasks, explicitly focusing on visual recognition and geometric recognition, respectively.
Finally, Section~\ref{Exploration_experiments} presents an ablation study to explore the fundamental components of FSVGP.

\subsection{Experimental setting}
\label{Experimental_setting}
\noindent{\textbf{Pre-trainig.}}
We use VG-FractalDB-1k (1000 categories, 1000 instances per category), ensuring equitable comparison with the existing FDSL methods. 
Following the approach of previous SSL and FDSL methods, we conduct pre-training using a unified transformer (Base)  for image recognition and a unified transformer (Small) for 3D object recognition. The Warm-up Cosine Scheduler is employed for scheduling during pre-training. The batch size is 64 for each GPU, the initial learning rate is 5e-4, the momentum is 0.9, the weight decay is 5e-2, and the number of epochs is 200. For example, training on VG-FractalDB-1k uses 16 NVIDIA V100 GPUs and requires about 60 hours.

\noindent{\textbf{Comparison methods.}}
For image recognition, we evaluate FSVGP against transformer-based pre-training methods, including supervised pre-training on ImageNet and SSL methods such as MAE~\cite{he2022masked} and DINO~\cite{Caron_2021_ICCV}. Additionally, we compare with FDSL methods, including ExFractalDB-21k~\cite{Kataoka_2022_CVPR}, RCDB-21k~\cite{Kataoka_2022_CVPR}, and VisualAtom-21k~\cite{takashima2023visual}. SAM~\cite{kirillov2023segany} is also included in comparisons for image detection and segmentation. 
For 3D object recognition, we focus on transformer-based pre-training methods suited to point clouds, comparing SSL approaches such as PointBERT~\cite{Yu_2022_CVPR}, PointMAE~\cite{pang2022masked}, and MaskPoint~\cite{liu2022masked} to evaluate FSVGP's effectiveness. For 3D object recognition, we specifically compare with the latest FDSL method, such as PC-FractalDB-1k~\cite{yamada2022point}. However, since PC-FractalDB-1k proposed pre-trained on VoteNet~\cite{qi2019deep}, to ensure a fair comparison, we also pre-train PC-FractalDB-1k using 3DETR~\cite{misra2021end}. For fine-tuning in geometric classification and segmentation, we utilize the backbone network of the PC-FractalDB-1k pre-trained model on 3DETR.

\noindent{\textbf{Fine-tuning datasets and evaluation metrics.}}
Table~\ref{tab:datasets} describes the detailed settings of fine-tuning datasets used in the experimental section. For detailed information on each fine-tuning dataset, the hyperparameters employed in the pre-training and fine-tuning processes, and the comparison baselines, please refer to the \SM{Supplementary Material}.

\begin{table*}[t]
    \begin{center}
    \caption{Comparison of the latest FDSL methods in image and 3D object classification (CLS), detection (DET), and segmentation (SEG). The best score is shown in \textbf{bold}. 
    }
    \vspace{-10pt}
    \scalebox{0.99}{ 
    \begin{tabular}{l|cccccc} 
    \toprule[0.8pt]
        & \multicolumn{3}{c}{\color{myblue}{Visual recognition}} & \multicolumn{3}{c}{\color{mygreen}{Geometric recognition}} \\
        Pre-training dataset & \color{myblue}{CLS} & \color{myblue}{DET} & \color{myblue}{SEG} & \color{mygreen}{CLS} & \color{mygreen}{DET} & \color{mygreen}{SEG}   \\
         & Acc. & AP$_{50}$ & AP$_{50}$ & Acc. & mAP$_{25}$ & mIoU$_{cat}$  \\
        \midrule[0.5pt]
        VisualAtom-21k & 91.3 & 66.3 & 63.3 & \xmark & \xmark & \xmark \\ 
        PC-FractalDB-1k & \xmark & \xmark & \xmark & 83.3 & 63.0 & 83.7  \\ 
        \rowcolor[gray]{0.9} VG-FractalDB-1k & \textbf{92.0} & \textbf{68.3} & \textbf{65.6}& \textbf{83.7} & \textbf{63.7} & \textbf{84.1}  \\ 
        \bottomrule[0.8pt]
    \end{tabular} 
    } 
    \vspace{-20pt}
    \label{tab:comparison_sixtask}
    \end{center}
\end{table*}

\subsection{FSVGP effects on image and 3D object recognition}
\label{brief_result}
 In the beginning, we begin by presenting in Table~\ref{tab:comparison_sixtask} a comparison against the latest FDSL methods across all six tasks in both image and 3D object recognition classification (CLS), detection (DET) and segmentation (SEG). 
 In image recognition, CLS signifies fine-tuning accuracy on ImageNet100, while DET and SEG refer to fine-tuning performance on MS COCO for detection and segmentation tasks. In 3D object recognition, CLS represents fine-tuning accuracy in ScanObjectNN (PB-T50-RS). 
DET involves fine-tuning performance on ScanNet, and SEG involves fine-tuning performance on ShapeNet-parts.
 
Table~\ref{tab:comparison_sixtask} demonstrates FSVGP (VG-FractalDB-1k) performs equally or better than VisualAtom-21k and PC-FractalDB-1k in all tasks of classification, detection, and segmentation for image and 3D object recognition.
VisualAtom-21k or PC-FractalDB-1k pre-trained models are designed to fine-tune datasets of the same modality, and it is considered difficult to apply them to datasets of different modalities. However, our FSVGP performs better on both visual and geometric recognition.
This result shows FSVGP's capability to process image and 3D object recognition using a unified pre-trained model by supervised pre-training with the formula-supervised consistency label.

In addition, we investigate the pre-training effects of learning visual-geometric representations. We evaluate the performance of pre-trained models on either visual or geometric modality compared to pre-trained models on both modalities.
Table~\ref{tab:cross-modality_classification} shows the results of pre-trained models on both real (ImageNet, ShapeNet) and synthetic (VG-FractalDB-1k) datasets.
All pre-training utilized the same supervised learning to conduct a fair comparison.
FSVGP (VG-FractalDB-1k, V + G), which uses both modalities, improves the recognition performance in both modalities and achieves similar performances to ImageNet and ShapeNet in visual and geometric recognition, respectively, even though VG-FractalDB consists of synthetic data. These results indicate the effect of FSVGP, which bridges the modality gap over single-modality pre-trained models.

In the following sections, we show the detailed analyses of FSVGP and comparisons with other pre-training on real datasets, such as SSL.

\begin{table}[t]
    \begin{center}
    \caption{Comparison of performance on pre-training on either visual or geometric modality with FSVGP (VG-FractalDB-1k, visual (V) + geometric (G)).
    }
    \label{tab:cross-modality_classification}
    \vspace{-18pt}
    \scalebox{0.84}{
    \begin{tabular}{l|cc|ccccccc|c|cc|c}
    \toprule[0.8pt]
        Dataset & Modal & \#data & \color{myblue}{C10} & \color{myblue}{C100} & \color{myblue}{Cars} & \color{myblue}{Flowers} & \color{myblue}{VOC12} & \color{myblue}{P30} & \color{myblue}{IN100}  & \color{myblue}{Avg.} & \color{mygreen}{M40} & \color{mygreen}{SONN}  & \color{mygreen}{Avg.} \\
        \midrule[0.5pt]
        ImageNet & V & 1M & \textbf{99.0} & \textbf{89.6} & \textbf{81.9} & \textbf{99.1} & \textbf{86.5} & \textbf{82.1} & \textbf{93.1} & \textbf{90.2} &  92.2 & 82.0 & 87.1 \\
        ShapeNet & G & 50k & 82.1 & 65.4 & 8.25 & 74.8 & 53.1 & 79.3 & 79.7 & 63.2 & \textbf{92.7} & \textbf{83.3} & \textbf{88.0} \\
        \midrule[0.5pt]
        VG-FractalDB-1k & V & 1M & 98.0 & 84.3 & 88.7 & 99.5 & 82.7 & 80.9 & 91.2 & 89.3  & 92.7 & 83.3 & 88.0 \\
        VG-FractalDB-1k & G & 1M &  87.5 & 68.4 & 11.2 & 82.1 & 57.6 & 80.6 & 82.8 & 67.2 & 92.6 & 83.3 & 88.0 \\
        \rowcolor[gray]{0.9} VG-FractalDB-1k & V+G & 1M & \textbf{98.1} & \textbf{85.9} & \textbf{89.2} & \textbf{99.5} & \textbf{83.5} & \textbf{81.7} & \textbf{92.0} & \textbf{90.0}  & \textbf{92.9} & \textbf{83.7} & \textbf{88.3} \\
        \bottomrule[0.8pt]
    \end{tabular}
    }
    \end{center}
    \vspace{-25pt}
\end{table}

\subsection{Comparative analysis in image recognition}
\vspace{-5pt}
\label{2D_image_recognition}
\noindent\textbf{Image classification.}
Table~\ref{tab:comparison_sl_ssl_fdsl} compares the fine-tuning results with existing pre-training methods (SL, SSL, and FDSL) in image classification. 
Table~\ref{tab:comparison_sl_ssl_fdsl} shows that the FSVGP (VG-FractalDB-1k) improvement compared with training from scratch (random initialization) in all datasets.
In addition, the FSVGP (VG-FractalDB-1k) pre-trained model shows improvement compared with previous FDSL methods. In particular, FSVGP (VG-FractalDB-1k) improves performance by Avg. +0.3\% from VisualAtom-21k despite the number of pre-training data is 1/21. 
However, FSVGP did not surpass the fine-tuning performance of SL, DINO, and MAE. Nevertheless, given real dataset issues associated with copyright, privacy, and social bias, we demonstrate the benefits of FSVGP.

\begin{table*}[t]
    \begin{center}
    \caption{Comparison of the latest supervised learning (SL), SSL, and FDSL methods in 2D image classification. `Modal' indicates a modality with `V'isual and/or `G'eometric inputs. The best scores for each learning type are shown in \textbf{bold}. 
    }
    \vspace{-18pt}
    \scalebox{0.81}{
    \begin{tabular}{lccc|ccccccc|c} 
    \toprule[0.8pt]
        Method & \#Data & Modal & Supervision &\color{myblue}{C10} & \color{myblue}{C100} & \color{myblue}{Cars} & \color{myblue}{Flowers} & \color{myblue}{VOC12} & \color{myblue}{P30} & \color{myblue}{IN100} & Avg. \\
        \midrule[0.5pt]
        From scratch & -- & -- & -- & 78.3 & 57.7 & 16.1 & 77.1 & 64.8 & 75.7 & 73.2 & 63.3 \\ \midrule[0.5pt]
        ImageNet-1k & 1.2M & V & SL & 99.0 & 89.6 & 81.9 & 99.1 & 86.5 & 82.1 & 93.1 & 90.2 \\
        ImageNet-1k & 1.2M & V & SSL (DINO) & 98.9 & 88.9 & \textbf{92.5} & 99.6 & 89.4 & 82.3 & 93.2 & 92.1 \\
        ImageNet-1k & 1.2M & V & SSL (MAE) & \textbf{99.1} & \textbf{90.1} & 91.3 & \textbf{99.8} & \textbf{90.2} & \textbf{82.8} & \textbf{94.1} & \textbf{92.5} \\
        \midrule[0.5pt]
        ExFractalDB-21k~\cite{Kataoka_2022_CVPR} & 21M & V & FDSL & 97.8 & 85.2 & 88.1 & \textbf{99.5} & 82.7 & 81.6 & 90.1 & 89.3 \\
        RCDB-21k~\cite{Kataoka_2022_CVPR} & 21M & V & FDSL & 96.8 & 82.9 & 85.9 & 99.0 & 81.2 & 81.2 & 90.2 & 88.2 \\
        VisualAtom-21k~\cite{takashima2023visual} & 21M & V & FDSL & 97.7 & \textbf{86.7} & \textbf{89.2} & 99.0 & 82.4 & 81.6 & 91.3 & 89.7 \\
        \rowcolor[gray]{0.9} VG-FractalDB-1k & 1.0M & V + G & FDSL (FSVGP) & \textbf{98.1} & 85.9 & \textbf{89.2} & \textbf{99.5} & \textbf{83.5} & \textbf{81.7} & \textbf{92.0} & \textbf{90.0} \\
        \bottomrule[0.8pt]
    \end{tabular}
    }
    \vspace{-10pt}
    \label{tab:comparison_sl_ssl_fdsl}
    \end{center}
\end{table*}

\noindent\textbf{Object detection and instance segmentation.}
We compare FSVGP with existing pre-training models for the fine-tuning results on average precision (AP) to COCO object detection and instance segmentation in Table~\ref{tab:detection}. 
We employ ViT-B as the backbone network and use MaskR-CNN as the detection head, referring from ViTDeT. Table~\ref{tab:detection} reports that FSVGP has superior results to training from scratch (random initialization) for COCO object detection and instance segmentation. 
In comparison to VisualAtom-21k, FSVGP (VG-FractalDB-1k) provides 2.5\% and 2.0\% higher AP in visual detection and segmentation, respectively. 
FSVGP (VG-FractalDB-1k) is inferior to SAM and MAE. However, FSVGP (VG-FractalDB-1k) outperforms ImageNet supervised pre-training and DINO despite synthetic pre-training.
From these results, ViT has a lower inductive bias, but because FSVGP is trained with visual and geometric modalities, it has the potential to give ViT a stronger inductive bias regarding spatial information and object shape than ImageNet supervised pre-training.

\noindent\textbf{ImageNet-1k classification.}
Table~\ref{tab:comparison_imagenet1k} presents the fine-tuning accuracy on ImageNet-1k with different image resolutions ($224 \times 224$ or $384 \times 384$), compared with representative conventional approaches.
Table~\ref{tab:comparison_imagenet1k} reveals that FSVGP (VG-FractalDB-1k) performs comparably to the VisualAtom-21k pre-trained model in image resolution, respectively.
Moreover, it is noteworthy that the large-scale FSVGP (VG-FractalDB-21k; 21000 categories, 1000 instances per category) achieves a similar performance of JFT-300M~\cite{dosovitskiy2021an} pre-training (83.8\% vs. 84.2\%), even though FSVGP using about 1/14 of the pre-training data when fine-tuning with image resolutions of $384 \times 384$. We consider that the result is worthwhile because JFT-300M is a non-public dataset. However, VG-FractalDB is more transparent and has fewer copyright, privacy, and social bias issues.

\begin{table}[t]
\begin{minipage}{0.52\textwidth}
\caption{
Comparison of representative pre-trained models in image object detection and instance segmentation.  The best values for each learning type are in \textbf{bold}.}
\label{tab:detection}  
    \vspace{-10pt}
    \scalebox{0.65}{
    \begin{tabular}{lccc} \toprule[0.8pt]
        Method & \color{myblue}{COCO Det} & \color{myblue}{COCO Ins Seg} \\
         &  AP$_{50}$ / AP / AP$_{75}$ & AP$_{50}$ / AP / AP$_{75}$ \\ \midrule[0.5pt]
        From scratch  & 65.7 / 45.5 / 49.3 & 62.8 / 40.4 / 43.7 \\ \midrule[0.5pt]
        ImageNet-1k (SL) & 63.9 / 43.1 / 47.4 & 60.9 / 38.9 / 41.7 \\ 
        ImageNet-1k (DINO) & 65.0 / 44.6 / 48.8 & 62.3 / 39.9 / 42.8 \\
        ImageNet-1k (MAE) & \textbf{70.7} / \textbf{50.5} / \textbf{55.4} & 68.1 / 44.8 / \textbf{48.6} \\
        SAM-1B (SAM) & \textbf{70.7} / \textbf{50.5} / 55.3 & \textbf{68.4} / \textbf{45.0} / 48.5 \\
        \midrule[0.5pt]
        ExFractalDB-21k & 66.8 / 46.1 / 50.3 & 63.8 / 40.7 / 43.4 \\
        RCDB-21k & 64.5 / 44.1 / 48.1 & 61.7 / 39.1 / 41.5 \\
        VisualAtom-21k & 66.3 / 45.4 / 49.8 & 63.3 / 40.4 / 42.9 \\
        \rowcolor[gray]{0.9} VG-FractalDB-1k (FSVGP) & \textbf{68.3} / \textbf{47.9} / \textbf{51.6} & \textbf{65.6} / \textbf{42.4} / \textbf{45.3} \\
        \bottomrule[0.8pt]
    \end{tabular}
    }
\end{minipage}
\hfill
\begin{minipage}{0.45\textwidth}
    \centering
    \caption{Comparison of fine-tuning accuracy on ImageNet-1k. The best value for each image resolution is in \textbf{bold}.}
    \label{tab:comparison_imagenet1k}  
     \vspace{-10pt}
   \scalebox{0.58}{
    \begin{tabular}{lccc} 
        \toprule[0.8pt]
        Method & Res. & Image  & \color{myblue}{ImageNet-1k} \\
        \midrule[0.5pt]
        From scratch & 224$^2$ & Real  & 80.5 \\
        ImageNet-1k (DINO)~\cite{Caron_2021_ICCV} & 224$^2$ & Real  & 82.8 \\
        ImageNet-1k (MAE)~\cite{he2022masked} & 224$^2$ & Real  & \textbf{83.6} \\
        ExFractalDB-21k~\cite{Kataoka_2022_CVPR}  & 224$^2$  &  Synthetic  & 82.7 \\
        RCDB-21k~\cite{Kataoka_2022_CVPR}  & 224$^2$  &  Synthetic  & 82.4 \\
        VisualAtom-21k~\cite{takashima2023visual}  & 224$^2$  &  Synthetic  & 82.7 \\
        \rowcolor[gray]{0.9} VG-FractalDB-1k (FSVGP) & 224$^2$  &  Synthetic  & 82.7 \\
        \midrule[0.5pt]
        From scratch & 384$^2$ & Real  & 81.2 \\
        ImageNet-21k (SL)~\cite{dosovitskiy2021an} & 384$^2$ & Real  & 83.0 \\
        JFT-300M (\textit{Dosovitskiy et al.,}) ~\cite{dosovitskiy2021an} & 384$^2$ & Real & \textbf{84.2} \\ 
        VisualAtom-21k~\cite{takashima2023visual} & 384$^2$ & Synthetic & 83.7 \\
        \rowcolor[gray]{0.9} VG-FractalDB-1k (FSVGP) & 384$^2$  &  Synthetic  & 83.6 \\
        \rowcolor[gray]{0.9} VG-FractalDB-21k (FSVGP) & 384$^2$  &  Synthetic  & 83.8 \\
        \bottomrule[0.8pt]
    \end{tabular}
    }
\end{minipage}
\vspace{-15pt}
\end{table}

\subsection{Comparative analysis in geometric recognition}
\vspace{-5pt}
\label{3D_object_recognition}
\noindent\textbf{3D object classification.} 
In Table~\ref{tab:3dcls}, we evaluate the fine-tuning accuracy on ModelNet40 and three subsets of ScanObjectNN, namely \{OBJ-BG, OBJ-ONLY, PB-T50-RS\} in 3D object classification. 
Table~\ref{tab:3dcls} shows that FSVGP (VG-FractalDB-1k) yields more accurate performance than training from scratch (random initialization), similar to the experimental results in image classification. 
Furthermore, FSVGP (VG-FractalDB-1k) improves the average accuracy in fine-tuning performance by +0.4\% compared to the latest FDSL method (PC-FractalDB).
In contrast, FSVGP (VG-FractalDB-1k) has lower fine-tuning accuracy than ShapeNet self-supervised pre-training methods (PointBERT, PointMAE, and MaskPoint).
One reason for this result may be that there are many overlapping categories, such as chairs, desks, etc., in the pre-training dataset (ShapeNet) and the fine-tuning datasets (ModelNet40 and ScanOjectNN). However, the goal of FSVGP is not only to perform the highest result on a specific task but also to achieve a superior pre-training effect for various tasks.

\begin{table*}[t]
    \centering
\caption{
Comparison of the latest supervised learning (SL), SSL, and FDSL methods in 3D object classification. `Modal' indicates a modality with `V'isual and/or `G'eometric inputs. The best score for each learning type is in \textbf{bold}. 
}
    \vspace{-10pt}
 \scalebox{0.78}{
    \begin{tabular}{lccc|cc|c} \toprule[0.8pt]
    \hspace{-3pt} Method & \#Data & Modal & Supervision & \color{mygreen}{ModelNet40} &  \color{mygreen}{ScanObjectNN} & Avg. \\
         & & &  & & OBJ-BG  / OBJ-ONLY / PB-T50-RS \\
         \midrule[0.5pt]
        From scratch & -- & -- & -- & 92.1 & 86.6 \hspace{8pt} / \hspace{8pt} 86.9 \hspace{8pt} / \hspace{8pt} 81.1 & 86.7 \\  
         \midrule[0.5pt]
        ShapeNet & 50k & G & SSL (Point-BERT) & \textbf{93.1} & \textbf{90.5} \hspace{8pt} / \hspace{8pt} \textbf{89.5 }\hspace{8pt} / \hspace{8pt} 85.0 & \textbf{89.5} \\
        ShapeNet & 50k & G & SSL (Point-MAE) & \textbf{93.1} & 90.4 \hspace{8pt} / \hspace{8pt} 88.1 \hspace{8pt} / \hspace{8pt} \textbf{85.8} & 89.4 \\
        ShapeNet & 50k & G & SSL (MaskPoint) & 92.8 & 89.5 \hspace{8pt} / \hspace{8pt} 88.1 \hspace{8pt} / \hspace{8pt} 83.8 & 88.6 \\
         \midrule[0.5pt]
         PC-FractalDB-1k & 1.0M & G  & FDSL & 92.6 & 88.3 \hspace{8pt} / \hspace{8pt} 88.3 \hspace{8pt} / \hspace{8pt} 83.3 & 88.1 \\
     \rowcolor[gray]{0.9} VG-FractalDB-1k & 1.0M & V + G & FDSL (FSVGP) & \textbf{92.9} & \textbf{88.9} \hspace{8pt} / \hspace{8pt} \textbf{88.5 }\hspace{8pt} / \hspace{8pt} \textbf{83.7} & \textbf{88.5} \\
        \bottomrule[0.8pt]
    \end{tabular}
    \label{tab:3dcls}
    }
    \vspace{-10pt}
\end{table*}

\begin{table}[t]
\begin{minipage}{0.45\textwidth}
\centering
\caption{Comparison of few-shot learning. 
The best averaged accuracy for each learning type is in \textbf{bold}. 
}
\vspace{-8pt}
\scalebox{0.85}{
 \begin{tabular}{lcccc} \toprule[0.8pt]
Method &  \multicolumn{4}{c}{\color{mygreen}{ModelNet40 Classification}} \\
& \multicolumn{2}{c}{5-way} & \multicolumn{2}{c}{10-way} \\
& 10-shot & 20-shot & 10-shot & 20-shot \\
\midrule[0.5pt] 
Scratch & 94.8  & 96.2 & 91.2 & 92.8 \\
\midrule[0.5pt] 
PointBERT & 94.6 & 96.3 & 91.0 & 92.7 \\
PointMAE & \textbf{96.3} & \textbf{97.8} & \textbf{92.6} & \textbf{95.0} \\
MaskPoint & 95.0 & 97.2 & 91.4 & 93.4 \\
\midrule[0.5pt] 
   PC-FDB-1k & 95.6 & \textbf{96.9} & 91.4 & 93.2 \\
  \rowcolor[gray]{0.9} VG-FDB-1k & \textbf{96.4} & 96.8 & \textbf{92.2} & \textbf{93.3} \\
 \bottomrule[0.9pt]
 \label{tab:few-shot}
\end{tabular}
}
\end{minipage}
\hfill
\begin{minipage}{0.5\textwidth}
\caption{Comparison of pre-training methods in 3D object detection and parts segmentation. 
The best value for each learning type is in \textbf{bold}. 
}
    \vspace{-10pt}
    \scalebox{0.87}{
    \begin{tabular}{lccc} \toprule[0.8pt]
            Method &  \color{mygreen}{ScanNetV2 Det} & \color{mygreen}{ShapeNet PartsSeg} \\
         &  mAP$_{25}$ / mAP$_{50}$ & mIoU$_{cat}$ / mIoU$_{ins}$ \\
        \midrule[0.5pt]
        Scratch  & 62.7 \hspace{4pt}  / \hspace{4pt} 37.5 & 83.3  \hspace{8pt} /  \hspace{8pt} 85.4 \\
        \midrule[0.5pt]
        PointBERT &  61.0 \hspace{4pt} / \hspace{4pt} 38.3 & 84.1  \hspace{8pt}  /  \hspace{8pt} 86.0 \\
        PointMAE &  -- \hspace{4pt} / \hspace{4pt} -- &  84.1  \hspace{8pt} /  \hspace{8pt} \textbf{86.1} \\
        MaskPoint &  \textbf{63.4}  \hspace{4pt}  / \hspace{4pt}  \textbf{40.6} &  \textbf{84.4} \hspace{8pt} /  \hspace{8pt} 86.0 \\
        \midrule[0.5pt]
        PC-FDB-1k &  63.0  \hspace{4pt}  / \hspace{4pt}  \textbf{42.5} &  83.7 \hspace{8pt} /  \hspace{8pt} \textbf{85.7} \\
        \rowcolor[gray]{0.9} VG-FDB-1k &  \textbf{63.7}  \hspace{4pt}  / \hspace{4pt} 42.0 &  \textbf{84.1} \hspace{8pt} /  \hspace{8pt} \textbf{85.7} \\
        \bottomrule[0.8pt]
        \label{tab:3ddet}
\end{tabular}
}
\end{minipage}
\vspace{-20pt}
\end{table}

\noindent\textbf{Few-shot learning.} 
We compare and verify the performance of few-shot learning on ModelNet40 in Table~\ref{tab:few-shot}. For few-shot learning, we randomly sample \textit{K} categories from ModelNet40 and \textit{N} shots of training samples from each category, following the experimental setting in~\cite{pang2022masked}.
FSVGP (VG-FractalDB-1k) achieves inspiring performance, outperforming training from scratch (random initialization) by a large margin in all few-shot settings despite different domain ModelNet40. 
In addition, FSVGP (VG-FractalDB-1k) achieves equal or better performance improvement over PC-FractalDB-1k in all few-shot settings. This result shows the FSVGP can achieve few-shot learning.

\noindent\textbf{3D object detection.} 
Table~\ref{tab:3ddet} reports fine-tuning performance for 3D bounding box mAP in 3D object detection. We fine-tune 3DETR on ScanNet. Our FSVGP (VG-FractalDB-1k) performs better under mAP$_{25}$ and mAP$_{50}$ than previous SSL methods. 
The FSVGP (VG-FractalDB-1k) is 1.4 points higher than MaskPoint~\cite{liu2022masked}, the latest SSL method (42.0 vs 40.6, mAP$_{50}$). 
More significantly, FSVGP (VG-FractalDB-1k) outperforms the PC-FractalDB-1k pre-trained model~\cite{yamada2022point} by 0.7 points on mAP$_{25}$, but falls short by 0.5 points on mAP$_{50}$.
This result suggests that the potential of FSVGP works well when applied to backbone networks of 3DETR.

\noindent\textbf{Parts segmentation.} 
Table~\ref{tab:3ddet} also compares the performance of FSVGP (VG-FractalDB-1k) with existing pre-training methods for parts segmentation. We fine-tune PointT-S on ShapeNet-parts and evaluate the performance using the mIoU for all categories (mIoU$_{cat}$) and all instances (mIoU$_{ins}$). 
Table~\ref{tab:3ddet} shows that our FSVGP (VG-FractalDB-1k) improves results over training from scratch (random initialization), for example, by 0.8 points (84.1 vs. 83.3, mIoU$_{cat}$). In addition, even though SSL duplicates pre-training and fine-tuning data, our method achieved performance comparable to SSL.
These results suggest that FSVGP is even more effective for recognizing more detailed 3D object structures.

\subsection{Ablation study}
\vspace{-5pt}
\label{Exploration_experiments}
In this section, we conduct additional experiments to explore essential factors of FSVGP guiding visual-geometric representation learning.
Specifically, we investigate to answer the following questions: 
(i) Which is more effective, fractal point clouds or CAD models? in FSVGP (ii) Can other generation rules be effective in FSVGP? Moreover, (iii) what is the effect of the pre-training task for VG-FractalDB? 

\begin{table*}[t]
\begin{minipage}{0.32\textwidth}
    \centering
    \caption{ShapeNet vs. VG-FractalDB (VG-FDB) in visual geometric pre-training. 
    The datasets were assigned visual and geometric modalities.
    }
    \vspace{-8pt}
    \scalebox{0.9}{
    \begin{tabular}{lccc}
    \toprule[0.8pt]
        Method & \#data & \color{myblue}{IN100} & \color{mygreen}{M40} \\
        \midrule[0.5pt]
        ShapeNet & 50k & 87.3 & 92.7 \\
        VG-FDB & 50k & \textbf{87.9} & \textbf{92.8} \\
        \bottomrule[0.8pt]
    \label{tab:shapenet}
    \end{tabular}
    }
\end{minipage}
\hspace{0.3mm}
\begin{minipage}{0.28\textwidth}
  \centering
\caption{Effect of generation rules. We compared Perlin noise (VG-PN-1k) with fractal (VG-FDB-1k) in FSVGP.}
\vspace{-10pt}
\scalebox{0.9}{
 \begin{tabular}{lccc} \toprule[0.8pt]
Method & \color{myblue}{IN100} &  \color{mygreen}{M40} \\ \midrule[0.5pt]
VG-PN-1k & 90.7 & 92.6 \\
VG-FDB-1k & \textbf{92.0} & \textbf{92.9} \\
\bottomrule[0.8pt]
  \label{tab:perlin}
\end{tabular}
}
\end{minipage}
\hspace{0.3mm}
\begin{minipage}{0.37\textwidth}
\caption{
Effect of supervisions in SSL (MAE) and our FSVGP. In both supervision settings, we utilized VG-FractalDB-1k (VG-FDB-1k) with visual and geometric modalities.
}
\vspace{-10pt}
\scalebox{0.88}{
 \begin{tabular}{lcc} \toprule[0.8pt]
Method & \color{myblue}{IN100} &  \color{mygreen}{M40} \\ \midrule[0.5pt]
VG-FDB-1k (MAE) & 80.3 & 92.8 \\
VG-FDB-1k (FSVGP) & \textbf{92.0} & \textbf{92.9} \\
\bottomrule[0.8pt]
\label{tab:pt_task}
\end{tabular}
}
\end{minipage}
\vspace{-20pt}
\end{table*}

\noindent\textbf{(i) Which is more effective, fractal point clouds or CAD models in FSVGP?}
We investigate the pre-training effects in FSVGP using VG-FractalDB and an existing 3D dataset, ShapeNet. 
Under the same conditions as VG-FractalDB, we project point clouds from ShapeNet onto images to generate visual-geometric data. Subsequently, the generated visual-geometric data from ShapeNet undergoes pre-training under the same conditions as FSVGP. Please refer to the \SM{Supplementary Material} for instances of image and point cloud data generated from ShapeNet.
To equalize the number of instances for ShapeNet and pre-training data, VG-FractalDB undergoes random sampling of data, with 50 instances per category. 
Table~\ref{tab:shapenet} shows that the VG-FractalDB pre-trained model is more accurate than the ShapeNet pre-trained model in ImageNet100 and ModelNet40 despite the same number of data.

\noindent\textbf{(ii) Can other generation rules be effective in FSVGP?}
We verify the pre-training effect regarding which generation rules (fractal and Perlin noise) are more effective in FSVGP.
We extended Perling Noise, effective as a dataset generation function~\cite{inoue2021initialization,Kataoka_2022_WACV}, to point clouds and constructed a Visual Geometric Perlin Noise (VG-PN) dataset. Please consult the \SM{Supplementary Material} for additional details on the generation process of the VG-PN dataset and examples of both point clouds and images.
We pre-train VG-PN with the same config and fine-tune it for image and 3D object classification. Table~\ref{tab:perlin} shows that the VG-FractalDB outperforms the VG-PN in ImageNet100 and ModelNet40. 

\noindent\textbf{(iii) Effect of pre-training tasks for VG-FractalDB.}
We explore the supervision types of VG-FractalDB by comparing the formula-supervised consistency label with self-supervision adopted by MAE, a representative SSL method. The implementations of decoders for the SSL are the same as MAE~\cite{he2022masked} and PointMAE~\cite{pang2022masked}, and the decoders reconstruct masked patches of 2D fractal images and 3D point clouds for pre-training based on the self-supervision.
Table~\ref{tab:pt_task} shows that utilizing the formula-supervised consistency label is more effective than utilizing the self-supervision based on MAE in ImageNet100 and ModelNet40.
\section{Discussion and Conclusion}
\vspace{-10pt}
This paper proposes FSVGP, which blends visual and geometric representations to achieve image and 3D object recognition on a unified transform model. 
FSVGP automatically generates fractal images, fractal point clouds, and their formula-supervised consistency labels based on fractal geometry. 

In the beginning, contrary to previous visual-geometric representation learning transfer the one-way knowledge using visual and geometric modalities, we show that FSVGP effectively achieved the pre-training effects in both image and 3D object classification, detection, and segmentation.
Furthermore, Table~\ref{tab:shapenet} and Table~\ref{tab:perlin} show that VG-FractalDB pre-training is more effective than the pre-training of ShapeNet and VG-PN dataset. We consider that these results are attributed to VG-FractalDB being generated based on more parameters of generation function, allowing it to pre-train on geometric shapes that are even more complex than those in the ShapeNet and VG-PN dataset.
Finally, Table~\ref{tab:pt_task} shows the effectiveness of the pre-training by simple classification task for VG-FractalDB and suggests that FSVGP is sufficient to learn visual-geometric representation in VG-FractalDB pre-training.
These observations show that FSVGP effectively improves fine-tuning performance in image recognition and 3D object recognition despite synthetic pre-training. Furthermore, FSVGP can reduce real dataset issues such as copyright, personal information, and social bias. 

\noindent{\textbf{Technical limitations and future work.}}
FSVGP is a pre-training method using VG-FractalDB (\textit{i.e.,} synthetic data).  
A previous FDSL study~\cite{nakashima2023does} reported that when FDSL is fine-tuning, a certain amount of real data regarding the domain gap between real and synthetic data is necessary. Therefore, compared to MAE (ImageNet), FSVGP (VG-FractalDB) is more challenging to work well with linear probing (Please consult the \SM{Supplementary Material} for more details).
Designing an efficient fine-tuning approach using FSVGP will be essential. 
We consider it important to validate FSVGP in 3D shape retrieval and multi-modal recognition involving bird's-eye view images and point clouds for future applications such as autonomous driving and search systems.

\section{Acknowledgments}
This paper is based on results obtained from a project, JPNP20006, commissioned by the New Energy and Industrial Technology Development Organization (NEDO). 
A computational resource, AI Bridging Cloud Infrastructure (ABCI), provided by the National Institute of Advanced Industrial Science and Technology (AIST), was used. 
We want to thank Ryota Suzuki, Yoshihiro Fukuhara, Naoya Chiba, Ryo Nakamura, Kodai Nakashima, Sora Takashima, Risa Shinoda, Masatoshi Tateno, Go Ohtani and Ryu Tadokoro for their helpful comments in the research discussions.

\bibliographystyle{splncs04}
\bibliography{main}

\appendix
\newpage

\vspace{5pt}
\begin{center}
	\textbf{\Large Supplementary Material}
\end{center}
\vspace{7pt}

\renewcommand{\thefigure}{\Alph{figure}}
\renewcommand{\thetable}{\Alph{table}}
\renewcommand{\thesection}{\Alph{section}}
\renewcommand{\theequation}{\Alph{equation}}
\setcounter{section}{0}
\setcounter{table}{0}
\setcounter{equation}{0}
\setcounter{figure}{0}

\section{FSVGP details}
\label{supp:fsvgp}
This section describes the details of our Formula-Supervised Visual-Geometric Pre-training (FSVGP). Section~\ref{supp:VG-FractalDB} details the Visual Geometric Fractal Database (VG-FractalDB). Section~\ref{supp:pre-train_model} details a unified model for pre-training VG-FractalDB.

\subsection{VG-FractalDB construction details}
\label{supp:VG-FractalDB}
This section delineates the methodology employed in constructing the VG-FractalDB, focusing on using 3D Iterated Function Systems (3D-IFS)~\cite{barnsley2014fractals} and our dataset diversity and consistency between visual and geometric modalities. 

3D-IFS is a mathematical framework for generating fractal geometry. It is central to defining the categories and variations in VG-FractalDB. 
Formula-supervised consistency labels in VG-FractalDB are linked to the 3D-IFS parameters. 
In certain 3D-IFS parameter cases, the 3D fractal point cloud is concentrated in a part of the 3D space. 
Therefore, the quality of the 3D fractal point cloud is checked based on the variance threshold to exclude such 3D fractal point clouds.
Only the 3D fractal point clouds whose variance value exceeds the variance threshold value in all axes are defined as the categories of VG-FractalDB. The variance threshold ensures a wide variety of fractal shapes. 
For augmenting within each category, we used FractalNoiseMix proposed by Yamada et al.~\cite{yamada2022point}. This augmentation technique enriches the dataset with a broader range of fractal geometries by augmenting 3D fractal models by mixing other 3D fractal models.

The 3D fractal models are then projected onto 2D planes to generate fractal images. This process randomly selects a camera viewpoint in 3D space. A perspective projection transformation maps point clouds onto a 2D plane. This particular transformation is chosen to accurately maintain the relative size and shape of 3D objects in the 2D rendering. Each parameter must be defined to achieve a realistic projection, such as the viewing angle (focal length), aspect ratio, and near and far planes. We set the focal length to 45 degrees, the aspect ratio to 1.0, and the near and far planes to 1.0 and 100, respectively.
The camera viewpoint setting is also an integral part of the projection process. This involves determining the camera's position, the point it is looking at, and its upward direction. These elements are used to compute a view matrix, which transforms the 3D objects from the world coordinate system to the camera coordinate system. 

For each 3D fractal model, a corresponding fractal image is generated from a randomly selected viewpoint. This approach ensures that each pair of 3D fractal point clouds and fractal images uniquely represents a particular viewpoint.
The resulting VG-FractalDB provides a rich 2D-3D fractal data representation for classification pre-training.

\subsection{Pre-training transformer model details}
\label{supp:pre-train_model}
We designed a single transformer model for learning VG-FractalDB. Our transformer model is built upon the standard Vision Transformer (ViT)~\cite{dosovitskiy2021an} and Point Transformer (PointT)~\cite{zhao2021point} structure, comprising transformer blocks. 
Each block includes a multi-head self-attention mechanism and a Multi-Layer Perceptron (MLP) block integrated with LayerNorm for normalization. 

The property of our single transformer model is to process both fractal images and 3D fractal point clouds through different embedding procedures tailored to the nature of each data type. 
For images, the image is then divided into patches of size $16\times16$, with each patch undergoing a linear projection to transform it into an embedding. 
For point cloud data, we start by downsampling a point cloud to a specific number of points. The downsampling point cloud is then clustered using a K-nearest neighbor, ensuring that local geometries within the cloud are preserved. These clustered points are passed through an MLP, generating embeddings.

Our transformer model is designed to be simple, learning visual-geometric representation from VG-FractalDB. 
Using distinct embedding processes for different data types showcases our transformer model's flexibility and potential to adapt diverse downstream tasks.

\section{Experimental setting details}
\label{supp:imple}
This section describes the experimental setup in detail. First, Section~\ref{supp:pre-train} describes the training setup in FSVGP. Sections~\ref{supp:2d_image_recgnition} and Section~\ref{supp:3d_object_recgnition} describe the experimental setup for image recognition and 3D object recognition, respectively. Finally, Section~\ref{supp:ablation} explains in detail the setup of the ablation study.

\subsection{Pre-training}
\label{supp:pre-train}
Our experiments set the hyperparameters based on the Data-efficient image Transformers (DeiT) model~\cite{touvron2021training}, as detailed in Table~\ref{tab:pre-train_params}. 
The training scripts were adapted from previous studies~\cite{takashima2023visual}, providing a foundational framework for our approach. 

\begin{table}[t]
    \begin{center}
	\caption{Pre-training setting.}
    \label{tab:pre-train_params}
     \vspace{-10pt}
    \scalebox{0.99}{
    \begin{tabular}{lcc} 
        \toprule[0.8pt]
        Config & \multicolumn{2}{c}{Value} \\ 
        & VG-FractalDB-1k & VG-FractalDB-21k \\
        \midrule[0.5pt]
        Epochs & 200 & 100\\
        Batch Size & 1024 & 8192 \\
        Optimizer & AdamW & AdamW \\
        LR & 5e-4 & 5e-4 \\
        Weight Decay & 0.05 & 0.05 \\
        LR Scheduler & Cosine decay & Cosine decay \\
        Warmup Steps & 5k & 5k \\
        Resolution & 224$\times$224 & 224$\times$224  \\ 
        Label Smoothing & 0.1 & 0.1 \\
        Drop Path & 0.1 & 0.1 \\
        Rand Augment & 9 / 0.5 & 9 / 0.5 \\
        Mixup & 0.8 & 0.8  \\
        Cutmix & 1.0 & 1.0  \\
        Erasing & 0.25 & 0.25 \\
        \bottomrule[0.8pt]
    \end{tabular}
    }
    \vspace{-10pt}
    \end{center}
\end{table}

\subsection{Image recognition}
\label{supp:2d_image_recgnition}
\noindent{\textbf{Image classification.}}
Our experiments validated our results using the image classification dataset that previous studies evaluated. 
We compare the top-1 accuracy during fine-tuning in 300 epochs as an evaluation metric.
Hyper-parameters at additional learning are shown in Table~\ref{tab:2d_class_params}. These are the same conditions as in the previous experimental setup in FDSL~\cite{takashima2023visual}.

\noindent{\textbf{Image object detection and instance segmentation.}}
This experiment was validated at MS COCO2017 using the official ViTDet~\cite{li2022exploring} GitHub. We used the hyperparameters of ViTDet as they are. The specific hyperparameters for the fine-tuning are shown in Table~\ref{tab:2D_obj_params}.

\begin{table}[t]
\begin{minipage}{0.48\textwidth}
\caption{Image classification setting.} 
\label{tab:2d_class_params}
\vspace{-10pt}
\scalebox{0.75}{
\begin{tabular}{lcc} 
        \toprule[0.8pt]
        Config & \multicolumn{2}{c}{Value} \\ 
        & ImageNet-1k & Others \\
        \midrule[0.5pt]
        Epochs & 300 & 300\\
        Batch Size & 1024 & 1024 \\
        Optimizer & AdamW & AdamW \\
        LR & 5e-4 & 5e-4 \\
        Weight Decay & 0.05 & 0.05 \\
        LR Scheduler & Cosine decay & Cosine decay \\
        Warmup Steps & 5 (epoch) & 5 (epoch) \\
        Resolution & 224$\times$224 / 384$\times$384 & 224$\times$224 \\
        Label Smoothing & 0.1 & 0.1 \\
        Drop Path & 0.1 & 0.1 \\
        Rand Augment & 9 / 0.5 & 9 / 0.5 \\
        Mixup & 0.8 & 0.8  \\
        Cutmix & 1.0 & 1.0  \\
        Erasing & 0.25 & 0.25 \\
        \bottomrule[0.8pt]
\end{tabular}
}
\end{minipage}
\hfill
\begin{minipage}{0.48\textwidth}
	\caption{Image object detection and instance segmentation setting.} 
    \label{tab:2D_obj_params}
    \vspace{-10pt}
    \scalebox{0.93}{
    \begin{tabular}{lcc} 
        \toprule[0.8pt]
        Config & \multicolumn{2}{c}{Value} \\ 
        &  From Scratch & Pre-train \\
        \midrule[0.5pt]
        Epochs & 30 & 30\\
        Batch Size & 16 & 16  \\
        Optimizer & AdamW & AdamW \\
        LR & 1.6e-4 & 4e-1  \\
        Weight Decay & 0.2  & 0.1  \\
        Warmup Steps & 1k & 1k \\
        Resolution & 1024$\times$1024 & 1024$\times$1024 \\
        Drop Path & 0.1/0.4 & 0.1/0.4 \\
        Large Scale Jitter & [0.1, 2.0] & [0.1, 2.0]  \\
        Rand Flip & 0.5 & 0.5  \\
        \bottomrule[0.8pt]
    \end{tabular}
    }
\end{minipage}
\vspace{-20pt}
\end{table}

\subsection{3D object recognition}
\label{supp:3d_object_recgnition}
\noindent{\textbf{3D object classification.}}
We used ModelNet40 and ScanObjectNN.
The evaluation was conducted on ModelNet40 and three ScanObjectNN subsets: OBJ-BG (including object surroundings), OBJ-ONLY (objects without background), and PB-T50-RS (a challenging subset with translated, rotated, and scaled objects). 
We employed the AdamW optimizer for fine-tuning and adjusted over 300 epochs using a cosine decay schedule. 
Models were fine-tuned on point clouds with 1024 points for ModelNet40 and 2048 points for ScanObjectNN, and performance was measured using overall accuracy, focusing on the highest accuracy achieved within 300 epochs. The specific hyperparameters for the fine-tuning are shown in Table~\ref{tab:3D_cls_params}.

\noindent{\textbf{Few-shot learning.}}
We conducted experiments by selecting $K$ classes from the ModelNet40 dataset and sampling $N+20$ objects from each class. These classes formed the basis for $K$-way, $N$-shot training subsets, with $K$ and $N$ varying between \{5, 10\} and \{10, 20\}, respectively. 
We created ten different subsets for these experiments and evaluated the model's performance by computing the mean and standard deviation of the highest accuracy obtained across these subsets. The AdamW optimizer was used during fine-tuning, adjusting it according to a cosine decay schedule over 150 epochs. We fine-tuned the model on ModelNet40 using point clouds of 1024 points each.

\begin{table}[t]
\begin{minipage}{0.48\textwidth}
\caption{3D object classification setting.} 
\vspace{-10pt}
\scalebox{0.75}{
\begin{tabular}{lcc} 
       \toprule[0.8pt]
        Config & \multicolumn{2}{c}{Value} \\ 
        & VG-FractalDB & Others \\
        \midrule[0.5pt]
        Epochs & 300 & 300\\
        Batch Size & 32 & 32 \\
        Optimizer & AdamW & AdamW \\
        LR & 5e-4 & 5e-4 \\
        Weight Decay & 0.05 & 0.05 \\
        LR Scheduler & Cosine decay & Cosine decay \\
        Warmup Steps & 10 (epoch) & 10 (epoch) \\
        Num. of Points & 1024(M)/2048(S) & 1024(M)/2048(S) \\
        Num of Patches & 64 & 64 \\
        Patch Size & 32 & 32 \\
        Augmentation & ScaleAndTranslate & ScaleAndTranslate \\
        \bottomrule[0.8pt]
\label{tab:3D_cls_params}
\end{tabular}
}
\end{minipage}
\hfill
\begin{minipage}{0.48\textwidth}
	\caption{3D object detection and parts segmentation setting.} 
     \vspace{-10pt}
    \scalebox{0.68}{
    \begin{tabular}{lcc} 
        \toprule[0.8pt]
        Config & \multicolumn{2}{c}{Value} \\ 
        & ScanNet & ShapeNet-parts \\
        \midrule[0.5pt]
        Epochs & 1080 & 300\\
        Batch Size & 32 & 32 \\
        Optimizer & AdamW & AdamW \\
        LR & 4e-4 & 5e-4 \\
        Weight Decay & 0.1 & 0.05 \\
        LR Scheduler & Linear warmup & Cosine decay \\
        Warmup Steps & 20 (epoch) & 10 (epoch) \\
        Num. of Points & 40000 & 2048 \\
        Num of Query/Patches & 256 & 64 \\
        Patch Size & -- & 32 \\
        Augmentation & RandomCuboid & ScaleAndTranslate \\
        \bottomrule[0.8pt]
    \label{tab:3D_det_params}
\end{tabular}
}
\end{minipage}
\vspace{-20pt}
\end{table}

\noindent{\textbf{3D object detection.}}
In our 3D object detection experiment, the ScanNet was used as a benchmark. We adopted the 3DETR model to fine-tune our 3D object detection approach, using its PointT-Small backbone network. The hyperparameters were tuned to those used in the original 3DETR. Our evaluation metrics were based on mean average precision (mAP) at 25\% and 50\% intersection over union (IoU). The specific fine-tuning hyperparameters are shown in Table~\ref{tab:3D_det_params}.

\noindent{\textbf{Parts segmentatoin.}}
We employed the ShapeNetPart dataset to evaluate part segmentation, which involves identifying detailed class labels for each point of a 3D model.  We assessed performance using the mean IoU (mIoU$_{ins}$) across all instances and IoU for each category. Furthermore, we reported the Mean IoU across all categories (mIoU$_{cat}$), ensuring equal treatment of each category in the dataset, irrespective of its frequency. This approach provides a comprehensive overview of the model's segmentation performance.
The specific hyperparameters for the fine-tuning are shown in Table~\ref{tab:3D_det_params}.

\begin{figure}[t]
  \centering
  \includegraphics[width=1.0\textwidth]{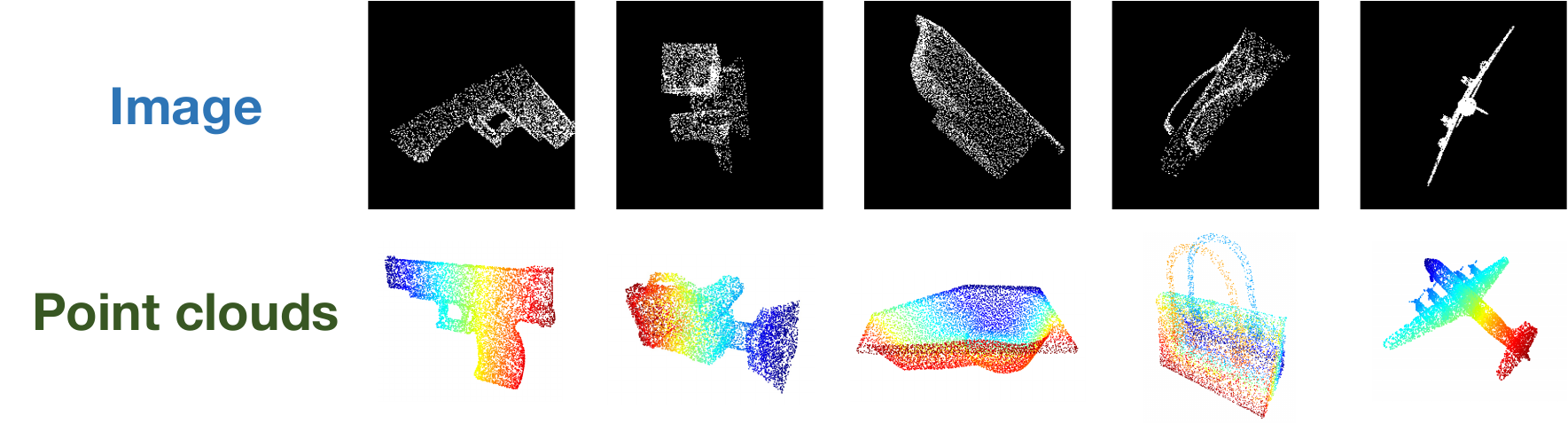}
  \caption{The examples of image and point cloud pair data in ShapeNet. }
  \label{fig:shapenet}
\end{figure}

\begin{figure}[t]
  \centering
  \includegraphics[width=1.0\textwidth]{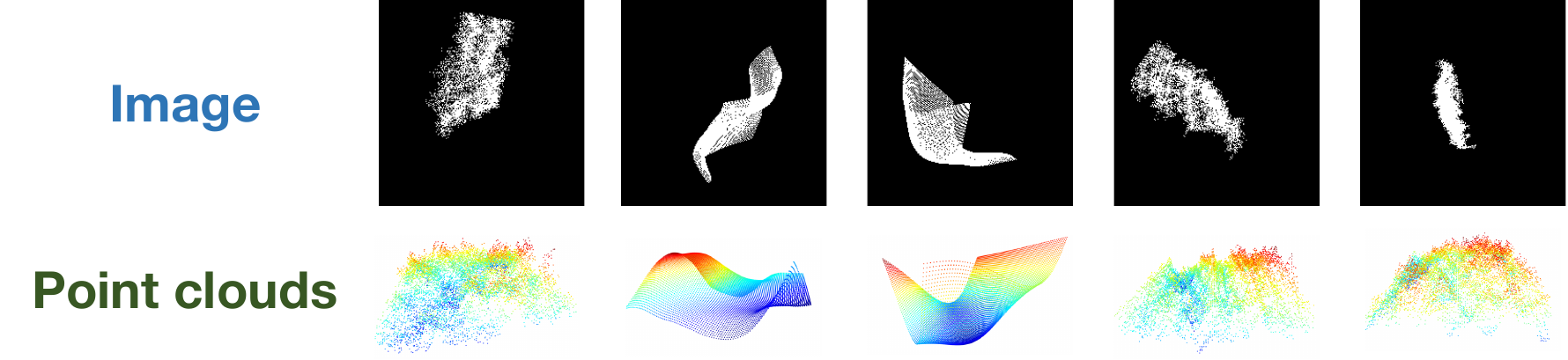}
  \caption{The examples of image and point cloud pair data in the Visual-Geometric Perlin Noise dataset. }
  \label{fig:pcpn}
\end{figure}

\subsection{Ablation study}
\label{supp:ablation}
\noindent\textbf{(i) Which is more effective, fractal point clouds or CAD models in FSVGP?}
In this experiment, we tested the pre-training effect of FSVGP by applying it to an existing 3D dataset, ShapeNet.
We generated images and point clouds for ShapeNet based on the VG-FractalDB construction procedure. Specifically, we project each 3D model of a ShapeNet onto an image from a random viewpoint position. An example of a generated image and point cloud is shown in Figure~\ref{fig:shapenet}.

\noindent\textbf{(ii) Can other generation rules be effective in FSVGP?}
In this experiment, we verified the pre-training effect of the generation rules by comparing fractal and Perlin noise in terms of the mathematical formula regularity that generates the data. Perlin noise is a gradient noise function for generating natural-looking textures and shapes, and previous studies~\cite{inoue2021initialization,Kataoka_2022_WACV} have reported its effectiveness in generating pre-trained datasets for image and video recognition. Therefore, we employed Perlin noise as the generating function to be compared in this experiment, considering its extensibility to 3D models.

We first generate 2D Perlin noise. Next, we lift the 2D Perlin noise to a point cloud. We then construct the Visual-Geometric Perlin Noise (VG-PN) dataset by projecting the point cloud onto an image. The 2D Perlin noise is pre-defined as a $100 \times 100$ grid. Random coordinates are determined within each grid, and a gradient vector is generated from the vertices of each grid based on these coordinates. The values in the grid are determined by linearly complementing the gradient vectors. The key parameters for generating the Perlin noise, the frequency, and scale, are varied within a specific range to ensure the diversity of the shape of the 3D Perlin noise. The VG-PN dataset defines these parameters as categories. The 2D Perlin noise is converted to a 3D Perlin noise as a point cloud by taking the values of each grid of the 2D Perlin noise as the Z-coordinate, finally, by projecting the 3D Perlin noise onto a image under the same conditions as VG-FractalDB. Finally, the 3D Perlin noise is projected onto the image under the same conditions as VG-FractalDB to generate the image/point cloud pair data, as shown in Figure~\ref{fig:pcpn}.

\begin{table}[t]
\begin{minipage}{0.48\textwidth}
\caption{Effect of formula supervision.}
\centering
\vspace{-10pt}
\scalebox{0.95}{
 \begin{tabular}{lcccc} \toprule[0.8pt]
Shuffle type & \color{myblue}{CIFAR100} &  \color{mygreen}{ModelNe40} \\ \midrule[0.5pt]
w/o shuffle & \textbf{85.9} & \textbf{92.9} \\
category & 84.4 & 92.7 \\
instance + category & 83.5 & 92.5 \\
 \bottomrule[0.8pt]
\label{tab:train_data}
\end{tabular}
}
\end{minipage}
\hfill
\begin{minipage}{0.48\textwidth}
  \caption{Effect of loss functions.}
  \centering
   \vspace{-10pt}
\scalebox{0.95}{
 \begin{tabular}{lcccc} \toprule[0.8pt]
Loss function & \color{myblue}{CIFAR100} &  \color{mygreen}{ModelNet40} \\ \midrule[0.5pt]
CE   & \textbf{85.9} & \textbf{92.9} \\
VGC  & 8.4 & 92.5 \\
CE + VGC & 85.4 & 92.2 \\
 \bottomrule[0.8pt]
\label{tab:effect_of_loss}
\end{tabular}
}
\end{minipage}
\vspace{-20pt}
\end{table}

\section{Additional experiments}
\subsection{What is the pre-training effect of collapsing the pair labels in VG-FractalDB?}
This experiment verifies the pre-training effect based on the formula-supervised consistency label.
We shuffled each pair of fractal data in VG-FractalDB to make it inconsistent.
Specifically, we implement two shuffle methods, named \textit{category} and \textit{instance + category}, which shuffle the categories of 3D fractal point clouds for each category and instance, respectively.
Let \(\mathbf{I} = \{\mathbf{I}^1, \mathbf{I}^2, \dots, \mathbf{I}^C\}\) and \(\mathbf{X} = \{\mathbf{X}^1, \mathbf{X}^2, \dots, \mathbf{X}^C\}\) denote the image and pointcloud data, respectively, where \(C\) is the number of categories.
The instances of images and point clouds in category \(c\) are denoted as \(\mathbf{I}^c = \{\mathbf{I}^c_1, \mathbf{I}^c_2, \dots, \mathbf{I}^c_M\}\) and \(\mathbf{X}^c = \{\mathbf{X}^c_1, \mathbf{X}^c_2, \dots, \mathbf{X}^c_M\}\), respectively, where \(M\) is the number of instances in each category.

The \textit{category} shuffle randomizes the category indices of point clouds to destroy the consistency of categories for images and point clouds.
After \textit{category} shuffle, the instances of point clouds in category \(c\) are denoted as \(\mathbf{X}^c_\mathrm{cs} = \{\mathbf{X}^{c'}_1, \mathbf{X}^{c'}_2, \dots, \mathbf{X}^{c'}_M\}\), where \(c'\) is the shuffled category index.
Therefore, the category labels for point cloud data are different from those for image data in the pre-training step, though the labels in each category are consistent for both images and point clouds.

The \textit{instance + category} shuffle randomizes both instance and category indices of point clouds to disrupt the consistency of instances for images and point clouds.
After \textit{instance + category} shuffle, the instances of point clouds in category \(c\) are denoted as \(\mathbf{X}^c_\mathrm{ics} = \{\mathbf{X}^{c'_1}_{i'_1}, \mathbf{X}^{c'_2}_{i'_2}, \dots, \mathbf{X}^{c'_M}_{i'_M}\}\), where \(c'_j\) and \(i'_j\) are the shuffled category and instance indices for the \(j\)-th instance, respectively.
Therefore, even the labels in each category are not consistent in point clouds.
Note that the shuffling methods exclusively randomize the labels for point cloud data to disrupt the consistency between image and point cloud data.
In other words, the labels associated with image data remain unaffected by the shuffling.

Table~\ref{tab:train_data} shows that FSVGP without shuffling was more effective than \textit{category shuffle} and \textit{instance + category} shuffle in CIFAR100 and ModelNet40.
This result shows that the formula-supervised consistency labels used in FSVGP improve the performance of pre-training.
The pre-training using the data shuffled by \textit{instance + category} still achieved reasonable results.
We believe pre-training on such data optimizes the model towards near-optimal parameters based on consistent image data, even though the shuffled point cloud data may impede convergence.
To validate the hypothesis, we examined the loss values both with and without the shuffling.
The values for image data were similar (2.45 vs 2.48), whereas the values for point cloud data differed significantly (6.90 vs 1.02).
In addition, the shuffling of both visual and geometric modalities disrupted the pre-training, causing a divergence in the loss values.

\subsection{Does the standard cross-entropy loss function alone suffice for pre-training in FSVGP?}
We contrast two scenarios: one employing CE loss based on the formula-supervised consistency label and another employing cross-entropy loss with a constraint term derived from visual-geometric correspondence (VGC).
We developed VGC as consistency labels, representing whether the pair of images and point cloud represent the same instance.
We shuffle the point cloud data instances in each category to generate a non-consistent pair. In each epoch of pre-training, we utilize both non-shuffled and shuffled data equally, randomly splitting the dataset in half.
VGC calculates the loss values using cross-entropy loss with consistency labels.
Table~\ref{tab:effect_of_loss} shows that FSVGP with only CE loss is better than the fine-tuning accuracy with VGC + CE loss.
This result finds that FSVGP learns visual-geometric representation with only CE loss rather than explicit visual-geometric correspondence terms such as VGC.

\subsection{Evaluation of the performance of pre-training models by linear probing}
Our experiment of this paper basically followed the evaluation protocols of previous FDSL studies. However, we believe that it is important to know about the feature representation that the pre-trained models learn through linear probing.
Therefore, we investigate the feature representations learned by FSVGP (VG-FractalDB-1k) and MAE (ImageNet). Specifically, we stop the gradient update of some transformer blocks in ViT during fine-tuning and evaluate which transformer block feature representations in ViT contribute to fine-tuning.

\begin{figure}[t]
  \centering
  \includegraphics[width=0.9\textwidth]{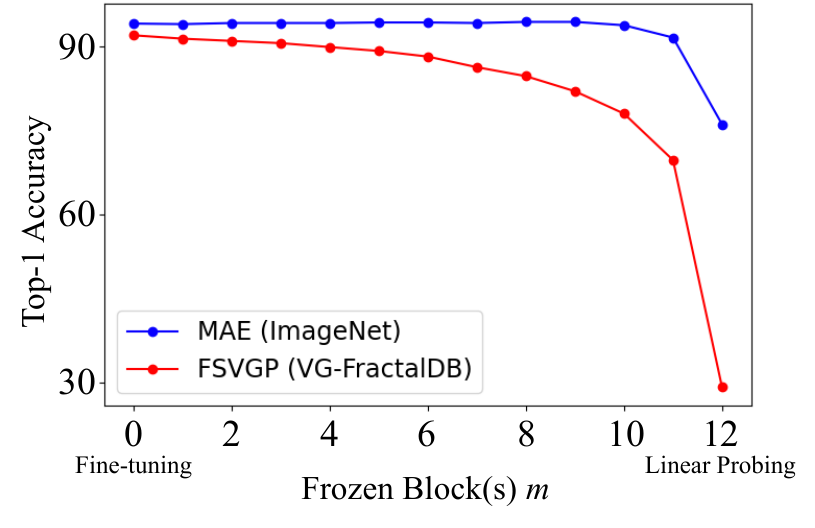}
  \caption{Comparison of classification accuracy when parameter update of each transformer block is frozen during fine-tuning of SVGP (VG-FractalDB-1k) and MAE (ImageNet). We use ViT-B on ImageNet100. }
  \label{fig:linear_prob}
  \vspace{10pt}
\end{figure}

We froze the first \textit{m} blocks of ViT-B during the fine-tuning (\textit{m} = 0 and 12 indicate full fine-tuning and linear probing, respectively).
As shown in Figure~\ref{fig:linear_prob}, although the difference in data domain between real images and fractal data degenerates the performance of FSVGP in linear probing, the fine-tuning from pre-trained representations significantly improves the performance. This result indicates the meaningful representation learned from FSVGP, especially in early layers.

\subsection{Multi-modal evaluations in 3D object classification}
We consider multi-modal evaluation important for showing the use case of FSVGP. Therefore, We conducted an initial experiment of 3D object classification using images and point clouds on ModelNet40. We confirmed that VG-FractalDB (V + G) outperforms VG-FractalDB (V or G) by +0.2 points and +0.6 points, respectively, when fine-tuning images and point clouds on ModelNet40. This result suggests the potential applications of FSVGP, such as autonomous driving with point clouds and bird's-eye view images.

\begin{figure}[t]
  \centering
  \includegraphics[width=1.0\textwidth]{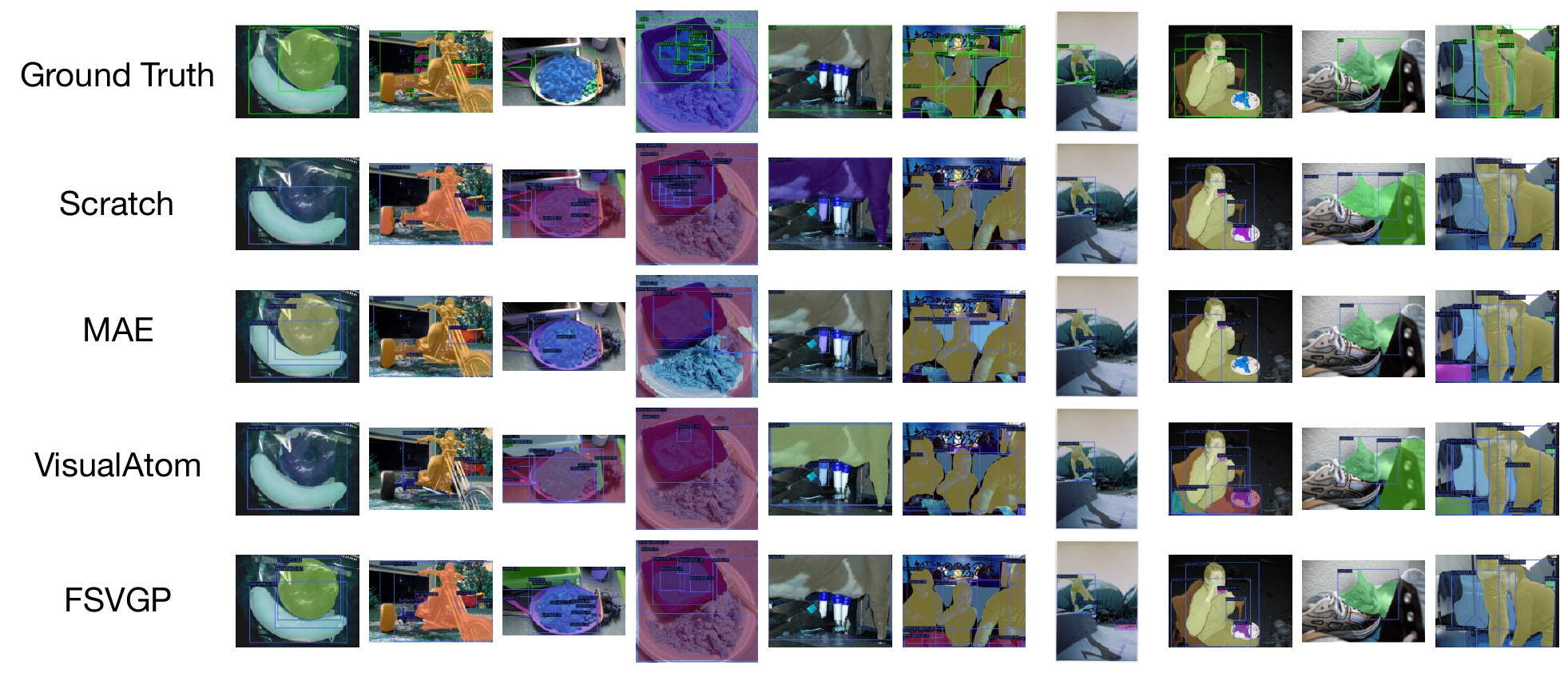}
  \caption{\textbf{FSVGP Success Cases:} compare ground truth with training from scratch, MAE, VisualAtom, and FSVGP output results.
We use VitDet (ViT-B) on MS COCO 2017 Val. }
  \label{fig:success_vis}
\end{figure}

\section{Qualitative examples}
\label{supp:qualitative_results}
The visualized predictions of the MS COCO underscore the ability of our FSVGP model to identify and delineate objects with high accuracy in complex scenes. Figure~\ref{fig:success_vis} demonstrates the FSVGP's accuracy in pinpointing object locations and discriminating between overlapping entities in detail-rich images.
For example, in Figure~\ref{fig:success_vis}, one can observe the FSVGP's acute precision in detecting and separating a cluster of beans on a plate, demonstrating its ability to locate and distinguish even the smallest objects. In addition, the figure highlights the model's ability to detect overlapping objects, such as a book partially obscured by a houseplant, demonstrating the nuanced recognition capabilities of FSVGP across a wide range of object categories.

\end{document}